\documentclass[lettersize,journal,twoside]{IEEEtran}
\usepackage{amsmath,amsfonts}
\usepackage{algorithmicx}
\usepackage{algorithm}
\usepackage{algpseudocode}
\usepackage{array}
\usepackage[caption=false,font=normalsize,labelfont=sf,textfont=sf]{subfig}
\usepackage{textcomp}
\usepackage{stfloats}
\usepackage{url}
\usepackage{verbatim}
\usepackage{graphicx}
\usepackage{cite}
\usepackage{color,soul,xcolor}
\usepackage{colortbl}
\usepackage{etoolbox}
\usepackage{makecell}
\usepackage{multirow}
\usepackage{threeparttable}
\usepackage{CJKutf8}
\usepackage{fancyhdr}
\soulregister{\cite}7 
\soulregister{\citep}7 
\soulregister{\citet}7 
\soulregister{\ref}7 
\soulregister\eqref7
\begin{document}

\title{Innovative Design of Multi-functional Supernumerary Robotic Limbs with Ellipsoid Workspace Optimization}

\author{Jun Huo, Jian Huang,~\IEEEmembership{Senior Member,~IEEE}, Jie Zuo, Bo Yang, Zhongzheng Fu, Xi Li\\ and Samer Mohammed,~\IEEEmembership{Senior Member,~IEEE}

\thanks{Manuscript received July 21, 2024; revised January 6, 2025; accepted July 8, 2025.
This work was supported in part by the National Natural Science Foundation of China under Grant U24A20280, 62333007, and U1913207, and in part by the Program for HUST Academic Frontier Youth Team. \textit{(Corresponding author: Jian Huang and Jie Zuo.)}

Jun Huo, Jian Huang, Zhongzheng Fu and Xi Li are with the Key Laboratory of the Ministry of Education for Image Processing and Intelligent Control, and the Hubei Key Laboratory of Brain-inspired Intelligent Systems, School of Artificial Intelligence and Automation, Huazhong University of Science and Technology, Wuhan 430074, China \texttt{huang\_jan@mail.hust.edu.cn}.

Jie Zuo is with the School of Information Engineering, Wuhan University of Technology, 122 Luoshi Road, Wuhan 430070, China \texttt{zuojie@whut.edu.cn}.

Bo Yang is with the State Key Laboratory of Intelligent Vehicle Safety Technology, Chongqing Changan Automobile Co Ltd, Chongqing 400023, China \texttt{ybandbob@gmail.com}.

Samer Mohammed is with the
Univ Paris-Est Créteil, LISSI, F-94400, Vitry, France \texttt{samer.mohammed@u-pec.fr}.
}
\thanks{
Digital Object Identifier (DOI): see top of this page.
}}

\pagestyle{fancy}
\fancyhf{}
\renewcommand{\headrulewidth}{0pt}
\fancyhead{} 
\fancyhead[RE]{\scriptsize IEEE TRANSACTIONS ON ROBOTICS. PREPRINT VERSION. ACCEPTED
JULY, 2025}
\fancyhead[LE]{\scriptsize \thepage}
\fancyhead[LO]{\footnotesize Huo $et$ $al$.: Innovative Design of Multi-functional Supernumerary Robotic Limbs with Ellipsoid Workspace Optimization}
\fancyhead[RO]{\scriptsize \thepage}
\markboth{IEEE TRANSACTIONS ON ROBOTICS. PREPRINT VERSION. ACCEPTED
JULY, 2025}%
{Shell \MakeLowercase{\textit{et al.}}: A Sample Article Using IEEEtran.cls for IEEE Journals}


\maketitle

\begin{abstract}
Supernumerary robotic limbs (SRLs) offer substantial potential in both the rehabilitation of hemiplegic patients and the enhancement of functional capabilities for healthy individuals. Designing a general-purpose SRL device is inherently challenging, particularly when developing a unified theoretical framework that meets the diverse functional requirements of both upper and lower limbs. In this paper, we propose a multi-objective optimization (MOO) design theory that integrates grasping workspace similarity, walking workspace similarity, braced force for sit-to-stand (STS) movements, and overall mass and inertia. A geometric vector quantification method is developed using an ellipsoid to represent the workspace, aiming to reduce computational complexity and address quantification challenges. The ellipsoid envelope transforms workspace points into ellipsoid attributes, providing a parametric description of the workspace. Furthermore, the STS static braced force assesses the effectiveness of force transmission. The overall mass and inertia restricts excessive link length. To facilitate rapid and stable convergence of the model to high-dimensional irregular Pareto fronts, we introduce a multi-subpopulation correction firefly algorithm. This algorithm incorporates a strategy involving attractive and repulsive domains to effectively handle the MOO task.
The optimized solution is utilized to redesign the prototype for experimentation to meet specified requirements. Six healthy participants and two hemiplegia patients participated in real experiments. Compared to the pre-optimization results, the average grasp success rate improved by 7.2\%, while the muscle activity during walking and STS tasks decreased by an average of 12.7\% and 25.1\%, respectively. The proposed design theory offers an efficient option for the design of multi-functional SRL mechanisms.
\end{abstract}


\section{Introduction}
\label{Intro}
\IEEEPARstart{W}{earable}
robots are used both to assist patients' rehabilitation and to augment the capabilities of healthy individuals.
Patients with hemiplegia require practical support and rehabilitation exercises to regain limb functionality and resume daily activities, thereby promoting self-assurance and mental well-being \cite{Lancet,Hasegawa2008}.
Such assistive robots enable patients to autonomously manage their activities of daily living, thereby reducing their reliance on caretakers and helping them to restore confidence \cite{Masahiro}. 
Additionally, there is a need to enhance the physical capabilities of healthy individuals through the integration of robotic devices that mimic human limbs and complement users' natural abilities. This integration can expand limb workspace, enhance grasping ability and hand flexibility, and prove hand strength for challenging tasks \cite{NMI2021}.
Consequently, whether aimed at rehabilitation for hemiplegia patients or functional enhancement for healthy individuals, the design of a supernumerary robotic limb (SRL) mechanism with general-purpose upper and lower limb functionalities plays a crucial role in improving the quality of assistance while reducing costs.

SRLs offer significant advantages and show significant potential in enhancing human capabilities \cite{BoTMRB,JAS-2020-0953}.
When the SRL mimics human upper limb functions, it can assist users in performing daily activities such as assisting people to hold objects \cite{Bonilla,FayeTRO,Irfan,Veronneau}, combing hair\cite{Domenico}, opening doors \cite{Guggenheim}, performing overhead operations \cite{LuoRAL}, and executing complex music performance \cite{Drum2018,Drum2021}. 
Parietti et al. proposed an SRL robot aimed at strengthening the lower limbs by providing assistance for standing and stabilizing the human body against the floor, walls, and surrounding structures \cite{PariettiTRO}.  
Khazoom et al. introduced a robotic assistive leg capable of rapidly swinging, adjusting to impact and providing substantial body support \cite{Khazoom}.
SRL can also support the human body in extreme environments, such as prone posture crawling \cite{Phillip} and squatting \cite{Parietti2014}. 
A common characteristic is the integration of additional heterogeneous limbs to reduce physical burden and enhance the functionality of human limbs.
Existing SRLs have largely achieved single-function assistance, while few studies focusing on multi-functional SRLs that incorporate general-purpose upper and lower limbs.
Meanwhile, there is lack of an applicable theory for designing general-purpose SRL.

To the best of our knowledge, there is currently no existing design theory guidance to address the balance between the flexibility required for upper limb functionality and the rigidity needed for lower limb support.
Therefore, it is challenging to establish an anthropomorphic general-purpose SRL design theory that adequately addresses the daily functional movements of human upper and lower limbs.
To address this gap, we propose a design theory for wearable robots, particularly an SRL aimed at assisting with grasping, walking, and sit-to-stand (STS) tasks. The design theory considers workspace similarity, static force requirements, and mechanism size.
Additionally, the mechanism's design should aim to minimize interference with the human natural limb to ensure smooth and natural interaction.
Furthermore, the method for quantifying the workspace and the algorithm framework for solving the design theory are introduced.

Wearable robots necessitate a high degree of alignment with the human body’s workspace. To achieve this goal, workspace similarity analysis is introduced.
A challenge arises in formulating a concise and efficient quantitative method to characterize workspace, due to the limitations of traditional point cloud workspace descriptions in capturing relevant properties.
Conventional point cloud spatial methods need traversing points within each workspace, leading to significant computational burdens and limitations in associating with geometric shapes.
Using ellipsoids to define workspace with specific parameters reduces computation complexity and increases processing efficiency, as they can approximate the crescent-shaped super-ellipsoidal nature found in actual workspace \cite{Kensuke,WU2020103711}. 
Ellipsoid methods are widely used in robotics research, such as operability ellipsoid \cite{MARIC2021103865}, maximum volume ellipsoid \cite{KARIMI201417}, inertia ellipsoid index \cite{Mostafa}, force ellipsoid \cite{IQBAL20141220}, and ellipsoid set membership filter \cite{WY_TASE}.
Maeda et al. introduced a concise representation of binary robotic arm workspace using an ellipsoidal outer approximation \cite{Kensuke}. Sorour et al. proposed a gripper kinematics model to generate point clouds for each finger's workspace \cite{Sorour}. Bayani et al. developed a convex optimization method to maximize gripper area and volume within the feasible workspace for planar and spatially indexed parallel robots \cite{Hassan2014}. Sirichotiyakul et al. presented a general framework to approximate the inner and exterior ellipsoids of singularity-free workspace for fully actuated robots \cite{Sirichotiyakul}.
Most of these approaches focus on describing workspace using ellipsoids. However, they do not extract ellipsoid-related features, which limits their ability to clarify the relationship between two ellipsoids. 
This study goes further by incorporating ellipsoid characteristics such as center distance, major axis, eccentricity, and volume. 
It lays the theoretical foundation for establishing evaluation criteria to assess workspace similarity between two ellipsoids.

To obtain precise and efficient computational results for mechanism design reference, a multi-objective optimization (MOO) requires the formulation of multiple sub-objectives. 
A MOO with several optimization sub-objectives will produce a high-dimensional irregular pareto front.
An inherent challenge in MOO is formulating sub-objective functions that enable rapid and reliable convergence to the high-dimensional irregular pareto front. 
Overcoming the challenge requires careful selection of objective model metrics. 
Current research have focused on MOO indicators, such as workspace \cite{CHAUDHURY2017115}, dimensions \cite{HuoMHS}, dexterity \cite{Ref11}, structural length \cite{Ref12}, and maximum singularity \cite{Ref13}. 
However, these indicators cannot be used for well describing the relationship between the anthropomorphic SRL workspace and the human limb workspace.
In addition, a combination of these indicators is insufficient as a optimization metric for multi-functional SRL.
Multiple indicators representing both upper limb flexibility and lower limb rigidity need to be carefully selected.
This paper integrates ellipsoidal workspace similarity, STS braced force, overall mass and inertia to formulate an indicator-based MOO model for determining the most appropriate link lengths.
Furthermore, the multi-object evolution algorithm (MOEA) is well-suited for solving problems involving high-dimensional irregular pareto front.
Two key factors to be carefully considered when utilizing a MOEA are: increasing global convergence towards the pareto front set, and improving individual enhancement probability \cite{Schutze}.
Traditional evolutionary algorithms, such as non-dominated sorting genetic algorithm (NSGA) \cite{NSGA}, particle swarm optimization \cite{MSPSO}, artificial bee colony, ant colony optimization, and firefly algorithm \cite{MEFA-CD}, cannot guarantee rapid convergence in MOO with high-dimensional irregular pareto fronts \cite{Hua2021}. 
When addressing MOO problems with high-dimensional irregular pareto fronts, multi-subpopulation and multi-direction algorithms show significant generalization \cite{YuTEVC}.
This study introduces an enhanced firefly algorithm called multi-subpopulation correction firefly algorithm (MSCFA), which integrates sub-population division and attraction-repulsion update techniques.

The main contributions of the paper are listed as follows. 
\begin{enumerate}
\item The optimization design theory for a wearable SRL robot, aimed at replicating the daily activities of human upper and lower limbs, is established. Based on the proposed theory, a prototype SRL robot with general-purpose upper and lower limb functions has been developed.
\item A geometric ellipsoid quantification method is proposed to represent the workspace, thereby reducing computational complexity and quantitatively extracting workspace properties. Additionally, a similarity metric is suggested to quantitatively describe the relationship between two ellipsoidal workspaces. 
\item An indicator based MOO model is developed. The model integrates the assessment of upper limb workspace similarity and lower limb workspace similarity in SRL robot. The proposed MSCFA algorithm is designed to achieve accurate and rapid solution convergence to high-dimensional irregular pareto front in MOO problem solving. 
\item The validity of the prototype is confirmed by both simulations and experiments. The prototype successfully achieved the goals of grasping with the upper limb, assisting in walking, and facilitating the transition from sit to stand.
\end{enumerate}

The rest of this paper is organized as follows.
In section II, the establishment of MOO model, description of the single objective function, and the solving algorithm are presented. Section III focuses on the simulation analysis. 
Section IV details the mechanical design, pilot study, and discussions. Section V summarizes the present study and offers potential directions for future works.

\section{Methods}
\label{Method}
\begin{figure*}[t!]
  \centering
  \includegraphics[width=0.8\textwidth]{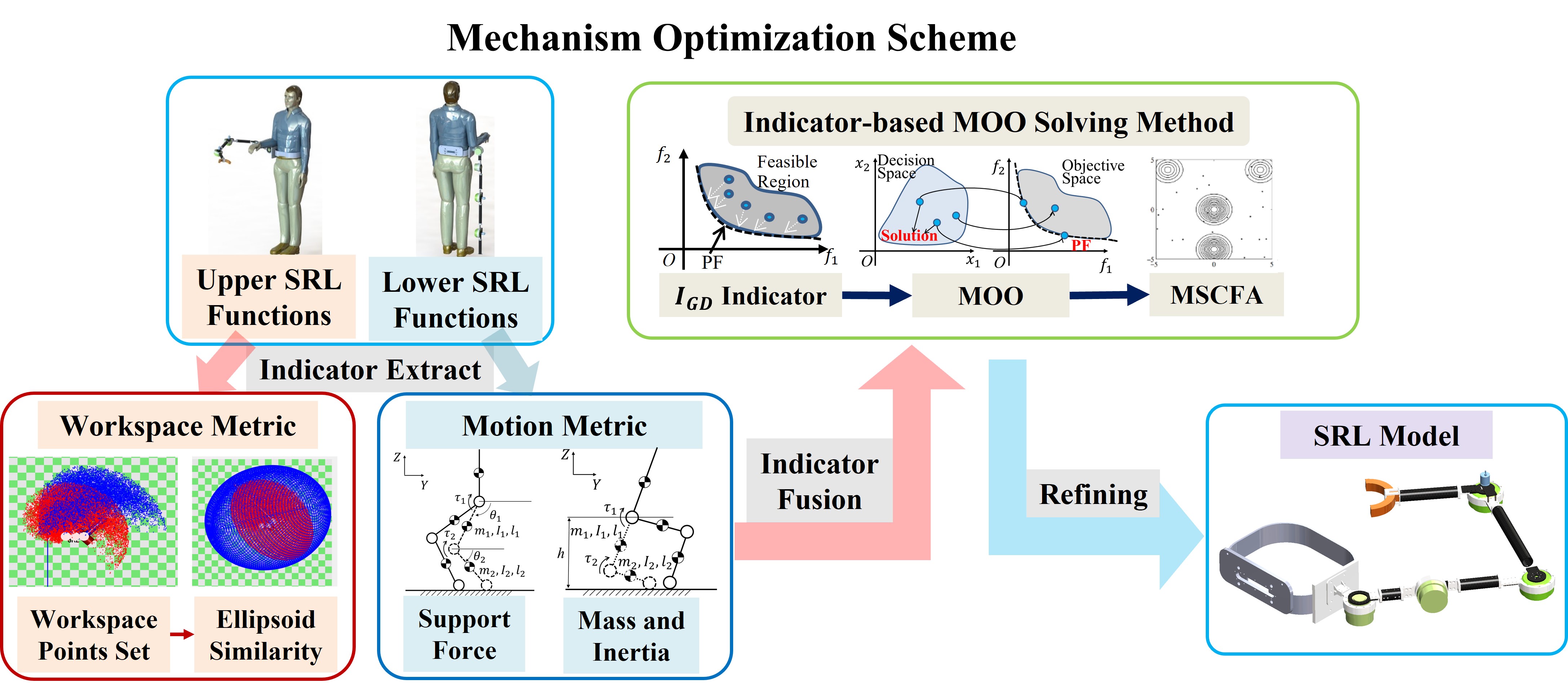}
  \caption{Block diagram overview of the system architecture of the general-purpose SRL robot.}
  \label{fig1}
\end{figure*}

In this section, the SRL mechanism and MOO model are introduced. 
The block diagram overview of the proposed framework is shown in Fig.~\ref{fig1}.
First, optimization indicators were extracted according to the required upper and lower limb functions: ellipsoid workspace similarity, STS static force, overall system mass and inertia. After the indicator fusion process, the MOO model is established and the MSCFA algorithm is proposed to solve this model. Finally, the obtained connecting link parameters are used to refine the mechanism.

\subsection{The Establishment of MOO Model}
\subsubsection{Degree of Freedoms (DoFs) Configuration for a General-purpose SRL Mechanism}
\begin{figure}[!t]
\centering
\includegraphics[width=0.95\columnwidth]{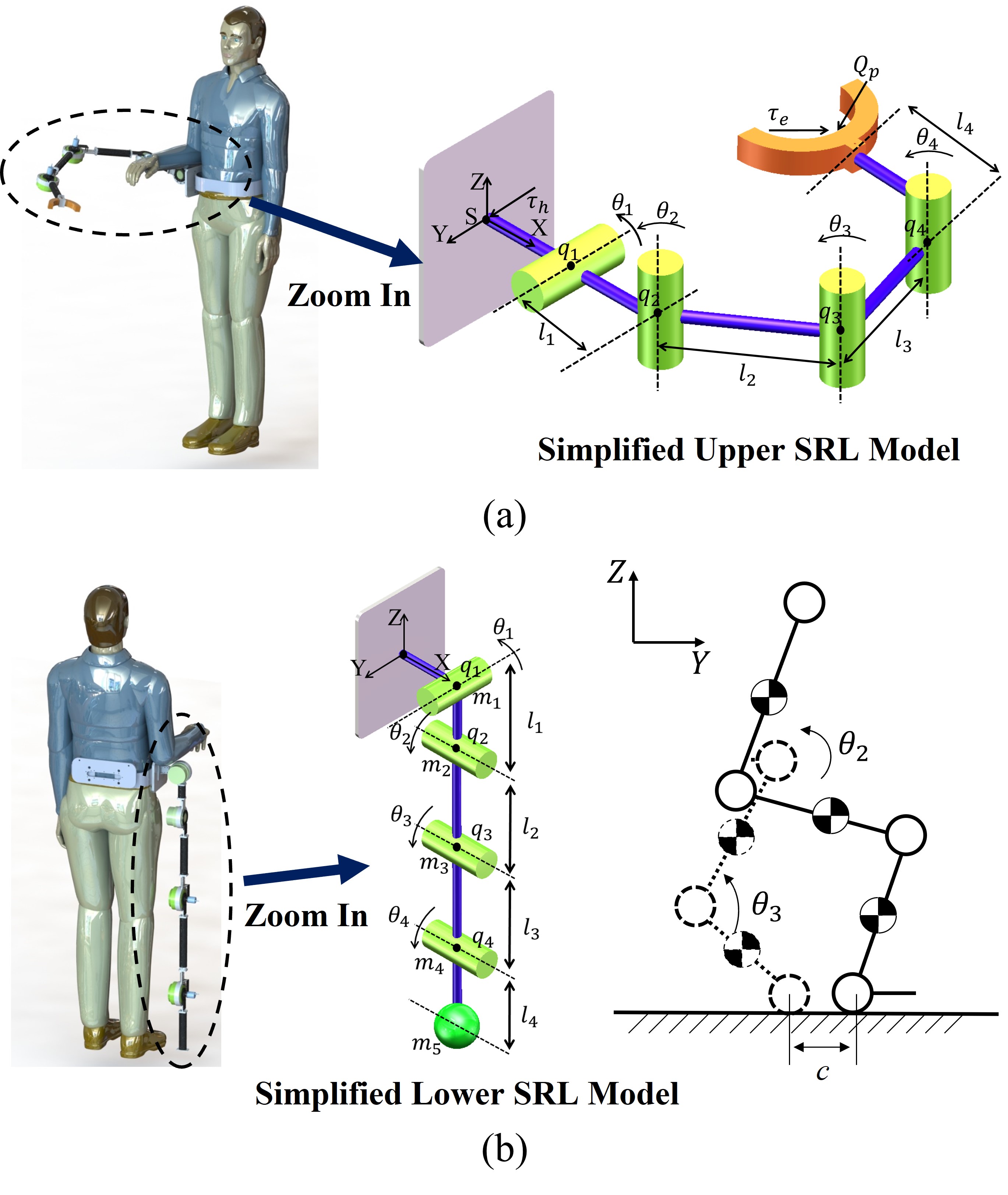}
\caption{The kinematics description of general-purpose SRL. (a) Case I, SRL kinematics description when using as an upper limb. (b) Case II, SRL kinematics description when using as a lower limb. The solid line links the human and the dashed lines the SRL.}
\label{Kine}
\end{figure}

The mechanism consists of a series-connected rod structure with 4 DoFs and an end effector, as shown in the simplified model in Fig.~\ref{Kine}. 
Where, $q_i$ denotes to the $i$-th joint, $\theta_i$ represents the rotation angle of the $i$-th joint, and $l_i$ is the $i$-th link length. The air pressure $Q_p$ is supplied to the soft gripper, which will be further detailed in the experimental section. Additionally, $\tau_h$ and $\tau_e$ refer to the forces exerted by the SRL on the human body and the environment, respectively. 
$m_i$ denotes the mass of the joint module (including motor and encoder etc.) and the end effector.
The parameter $c$ represents the horizontal distance from the robot’s pivot to the human’s footsteps during STS motion. 
It is important to note that the SRL is mounted at the waist, above the pelvis, on the side of the impaired limb.
The mechanism supports two distinct functionalities. The first DoF, located at the fixed support pivot, serves as a transitional link between the upper and lower limbs, adjusting posture as needed (see supplementary video for a demonstration of this switching action). When used as an upper limb, the end effector is a flexible gripper that bends and deforms when filled with compressed air \cite{ru2023JoVE}. When used as a lower limb, the end effector is replaced with a spherical rubber component with a high friction coefficient.
\subsubsection{The Dual Functionality Envisaged for the SRL Mechanism}
The challenge lies in creating a general-purpose SRL model that can effectively address both upper and lower limb functionality. Specifically, this requires: 

\textit{a. Reaching and grasping functionality:} When SRL is used to strengthen the upper limb for the healthy or replace the upper limb for the paralyzed (right arm in Fig.~\ref{Kine}(a)), the upper general-purpose SRL should be able to assist the user to perform daily functional activities such as reaching and grasping in standing or siting position.
In its upper limb function, all four DoFs are movable to ensure maximum workspace and dexterity.
Meanwhile, the pneumatic gripper contracts to grip objects when it reaches the intended area. 

\textit{b. Standing support functionality:} When SRL is used as a lower limb to assist the user’s walking or STS transfer (right side in Fig.~\ref{Kine}(b)), the lower general-purpose SRL should be able to provide reliable support during walking or STS transition. Clinical recommendations for using a conventional unilateral lower SRL suggest that it should be positioned 15 to 20 cm lateral to the user's toes on the paralyzed side for optimal support \cite{YanTMech}.
In lower limb function, $q_1$ and $q_4$ are locked, while $q_2$ and $q_3$ are utilized as the hip and knee, respectively, to perform walking and STS movements in YZ plane.
Traditional canes are handheld and involve upper-body engagement, whereas our SRL is entirely hands-free and body-mounted. The SRL segment extends downward from the waist and functions as a ground-contacting limb that supplements stability during gait. This is conceptually similar to the third-leg support paradigm established in previous SRL locomotion studies \cite{Khazoom}.

In the envisioned application scenarios, the current design primarily targets unilateral motion rather than bi-manual or tri-manual manipulation using the SRL.

\subsubsection{MOO Model}
In metric integration, the objective function method is commonly applied to optimize designs in the field of robotics \cite{KELAIAIA2012159,ZHOU2012113}. Among these approaches, the indicator-based MOEA stands out for its operational simplicity and strong interpretability \cite{Cardona2020}.
The Generational Distance Indicator ($I_{GD}$) reports how far, on average, $\mathcal A$ is from $\mathcal {PF}$, where $\mathcal {PF}$ is the true Pareto Front and $\mathcal A$ is an approximation of the true Pareto Front. $I_{GD}$ is Pareto non-compliant and it is defined as:
\begin{equation}
\label{pf}
I_{GD}(\boldsymbol D) = \frac{1}{|\mathcal A|}(\sum^{|\mathcal A|}_{i=1}d_i^p)^{\frac{1}{p}},
\end{equation}
where, $|\mathcal A|$ is the number of vectors in $\mathcal{A}$, $\boldsymbol D = [d_1, d_2, ..., d_i]^T$ is the Euclidean phenotype distance vector between each member and the $\mathcal A$. If $I_{GD}=0, \mathcal{A} \subseteq \mathcal{PF}$.
$p$ is the calculation order, and it is commonly defined $p=2$.

\begin{figure}[!t]
\centering
\includegraphics[width=2.5in]{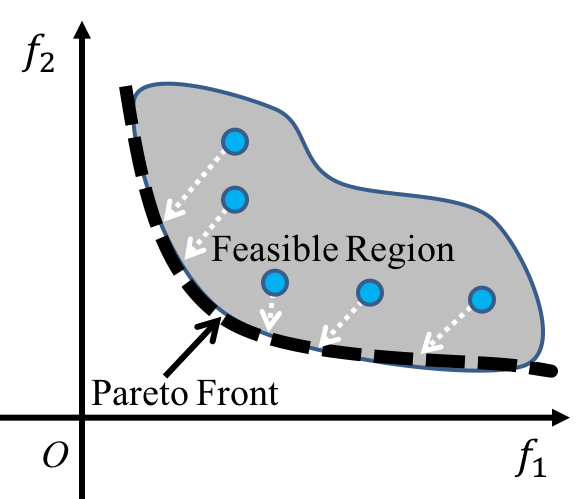}
\caption{Pareto Front of MOO problem under 2-dimension condition. The blue dots refer to the solving individuals of MOO problem.}
\label{PF}
\end{figure}

Without loss of generality, a MOO problem can be defined as follows,
\begin{equation}
\label{MOP}
\begin{aligned}
&{\rm min} \quad \boldsymbol F(\boldsymbol x) = (f_1(\boldsymbol x), f_2(\boldsymbol x), ..., f_M(\boldsymbol x))^T,\\
&{\rm s.t.} \quad \quad \mathbf x \in \Omega,
\end{aligned}
\end{equation}
where $\boldsymbol x=(x_1,x_2,...,x_D)^T$ is a D-dimensional decision vector, and $\Omega$ is the decision space. $\boldsymbol F(\boldsymbol x)$ denotes the objective vector with $M$ objectives.
In our problem formulation, the MOO problem can be expressed as:
\begin{equation}
\label{MOP2}
\begin{aligned}
&{\rm min} \quad \Phi(\boldsymbol{x}) = I_{GD}(\boldsymbol D(\boldsymbol x)),\\
&{\rm s.t.} \quad \quad \mathbf{LB} \leq \boldsymbol x \leq \mathbf{UB},
\end{aligned}
\end{equation}
where, $\Phi(\boldsymbol{x})$ is a 11-dimension objective function, $\boldsymbol D(\boldsymbol x)=[\boldsymbol D_1(W_{SRL},W_{h}), \boldsymbol D_2(W_{SRL},W_{c}), \boldsymbol D_3(\boldsymbol x), \boldsymbol D_4(\boldsymbol x)]^T$, $\boldsymbol x=(l_1,l_2, l_3, l_4, c)^T$ is a 5-dimension decision vector. 
$W_{SRL}$ refers to the workspace of SRL.
$W_h$ denotes the workspace for the endpoint of the human right hand.
$W_c$ represents the workspace for the endpoint of the cane when the right hand of the human body is using the cane for support \cite{HuoMHS}.
$\mathbf  {LB} = [0.1, 0.1, 0.1, 0.1, -0.5]^T$ and $\mathbf{UB}=[0.6, 0.6, 0.6, 0.6, 0.5]^T$ are the bound limits vector.
It should be noted that the theoretical lower bound of the link length is 0, but due to the minimum link length required in the actual engineering manufacturing process being 0.1 m, the $\mathbf{LB}$ of link length is set to 0.1 m. Considering the size of the human body, the length of a single link should not exceed half the length of the leg. The sitting fixed point is set 0.5 m before and after the human focal point based on practical experience.
\begin{equation}
\label{MOP3}
\begin{aligned}
&\boldsymbol D_1(W_{SRL},W_{h}) = [d_1,d_2,d_3,d_4]=[f_1(\boldsymbol x, W_h)-\mathcal{PF}_1, \\
&\quad \quad \quad \quad \quad \quad \quad f_2(\boldsymbol x, W_h)-\mathcal{PF}_2,f_3(\boldsymbol x, W_h)-\mathcal{PF}_3,\\ 
&\quad \quad \quad \quad \quad \quad \quad f_4(\boldsymbol x, W_h)-\mathcal{PF}_4],\\
&\boldsymbol D_2(W_{SRL},W_{c}) = [d_5,d_6,d_7,d_8]=[f_1(\boldsymbol x, W_c)-\mathcal{PF}_5, \\
&\quad \quad \quad \quad \quad \quad \quad f_2(\boldsymbol x, W_c)-\mathcal{PF}_6, f_3(\boldsymbol x, W_c)-\mathcal{PF}_7,\\
&\quad \quad \quad \quad \quad \quad \quad f_4(\boldsymbol x, W_c)-\mathcal{PF}_8],\\
&\boldsymbol D_3(\boldsymbol x) =  d_9=f_5(\boldsymbol x)-\mathcal{PF}_9,\\
&\boldsymbol D_4(\boldsymbol x) = [d_{10},d_{11}]=[f_6(\boldsymbol x)-\mathcal{PF}_{10},f_7(\boldsymbol x)-\mathcal{PF}_{11}],
\end{aligned}
\end{equation}
where, $\boldsymbol {\mathcal{PF}}$ is a 11-dimension optimal solution vector.
According to the objective function and constraints in Eq.~(\ref{MOP2}), 
$\boldsymbol D_i(\cdot)$ refers to the distance between the solution and the $\boldsymbol {\mathcal{PF}}$ of the MOO problem, shown in Fig.~\ref{PF}.
$f_i(\cdot)$ represents the single objective function of MOO, as defined in the subsequent section.

\subsection{Description of Sub-objective Function}
In this section, four sub-objective functions, including ellipsoid workspace similarity, STS static force, relative mass, and moment of inertia will be defined.
\subsubsection{Workspace Similarity}
To measure the similarity of two workspace point sets, it is necessary to calculate the two workspace point sets individually, obtain the minimum-volume enclosing ellipsoid for each workspace point set, and finally calculate the workspace similarity based on the properties of the ellipsoids.

\textit{Step 1, Workspace Points Set: }
The workspace of a robot is the set of reachable points at the end of the manipulator given all positions and poses. 
Common tools for describing rigid body kinematics of robots include Denavit–Hartenberg (D-H) parameters \cite{craig2005introduction} and screw theory \cite{murray2017mathematical}. D-H parameters are summarized in supplenmentary material.
Screw theory is used to define the workspace of the robot manipulator and human limbs.
The forward kinematics map, $g_{st}: Q\to \rm {SE(3)}$ is given by
\begin{equation}
\label{gst}
g_{st}(\boldsymbol \theta)= e^{\boldsymbol \xi_1\theta_1}e^{\boldsymbol \xi_2\theta_2}...e^{\boldsymbol \xi_n\theta_n}g_{st}(0),
\end{equation}
where, $\boldsymbol \xi_i$ is a twist which corresponds to the screw motion for the $i$-th joint with all other joint angles held fixed at $\theta_j=0$. $Q$ is a $\rm SE(3)$ subgroup of joint rotation angle.
For a revolute joint, the twist $\boldsymbol \xi_i$ has the form
\begin{equation}
\label{xi}
\boldsymbol \xi_i= 
\begin{bmatrix}
    -\boldsymbol \omega_i\times \boldsymbol q_i\\
    \boldsymbol \omega_i
\end{bmatrix},
\end{equation}
where $\boldsymbol \omega_i\in \mathbb{R}^3$ is a unit vector in the direction of the twist axis and $\boldsymbol q_i \in \mathbb{R}^3$ is any point on the axis.

The workspace of the manipulator is defined as the set of all end-effector configurations reachable by varying joint angles. If $Q$ is the configuration space of a manipulator and $g_{st}: Q\to \rm {SE(3)}$ is the forward kinematics map, then the workspace $W$ is defined as the set,
\begin{equation}
\label{Set1}
W=\{g_{st}(\boldsymbol \theta): \boldsymbol \theta \in Q\} \subset \rm {SE(3)}.
\end{equation}

In addition, the workspace of human hand $W_h$ and the workspace of using a cane $W_c$ are also needed in this task. The workspace of human limbs can be obtained by establishing kinematic models \cite{abdel2004}. 
The formula of workspace $W_h$ and $W_c$ are as follows,
\begin{equation}
\label{Set2}
W_h=\{g_{h}(\boldsymbol \theta): \boldsymbol \theta \in Q\} \subset \rm {SE(3)},
\end{equation}
\begin{equation}
\label{Set3}
W_c=\{g_{c}(\boldsymbol \theta): \boldsymbol \theta \in Q\} \subset \rm {SE(3)}.
\end{equation}

It should be noted that when the SRL contacts the ground, a closed loop linkage is formed, leading to a reduced workspace. The calculation of the reduced workspace is detailed in supplenmentary material.

\textit{Step 2, Minimum-Volume Enclosing Ellipsoid: }
The goal of this step is to find the minimal circumscribed ellipsoid that envelops all workspace areas, given a definite workspace representation.
The ellipsoid formulation in Euclidean space is as follows \cite{kumar2005minimum,bowman2023computing}, 
\begin{equation}
\label{E1}
\begin{aligned}
    {\rm min}& \quad \rm log(\rm det(\mathbf A)),\\
    {\rm s.t.}& \quad (\mathbf {p_i-c})^T \mathbf A (\mathbf{p_i-c})\leq 1,
\end{aligned}
\end{equation}
where, the vector $\mathbf{c}$ is the estimated ellipsoid center, $\mathbf{p_i}$ is the $i$-th point vector of the workspace dataset. The aim is to find the tri-axial minimum enveloping ellipsoid $\mathbf{A}$ to define the workspace.
Matrix $\mathbf{A}$ contains information about the size and shape of the ellipsoid.
Positive constant $r_1, r_2, r_3\ (r_1>r_2>r_3>0)$ are the semi major-axis, semi middle-axis and semi minor-axis of the ellipsoid, respectively, which can be derived by matrix $\mathbf A$.
\begin{equation}
\label{SVD}
[\mathbf{U\quad Q\quad V}] = \rm svd(\mathbf A),
\end{equation}
where, $\mathbf{U}$ and $\mathbf{V}$ are the rotating matrix, $\mathbf{Q}$ is the semi axis sector, which is $r_i=\frac{1}{\sqrt{Q(i,i)}}$.

\begin{figure}[!t]
\centering
\includegraphics[width=2.5in]{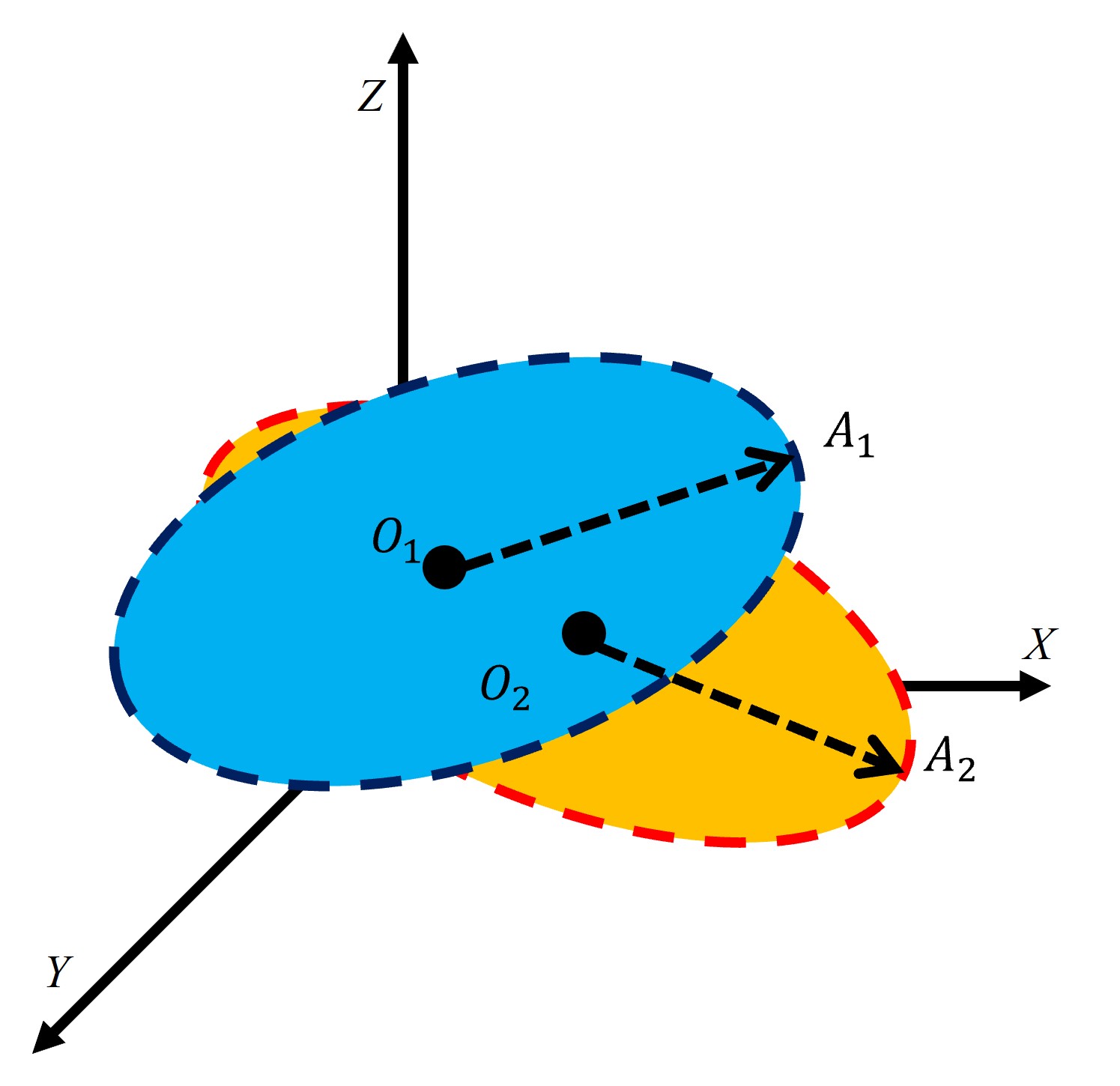}
\caption{The description of two workspace ellipsoid.}
\label{Epoi}
\end{figure}

\textit{Step 3, Workspace Similarity: }
We then define a similarity index of the two ellipsoids to describe the workspace similarity, as shown in Fig.~\ref{Epoi}.
Four indicators are used to define the ellipsoid similarity:  center distance to define position similarity ($O_1,O_2$ in Fig.~\ref{Epoi}), semi major-axis distance to define posture similarity ($\overrightarrow{O_1 A_1}, \overrightarrow{O_2 A_2}$ in Fig.~\ref{Epoi}), oblateness to define shape similarity, and volume to define size similarity. 

\begin{itemize}
\item \textit{Center Distance (CD):}
CD represents the distance between the centers of two ellipsoids.
The distance of center point $O_1, O_2$ is given by
\begin{equation}
\label{F1}
\quad f_1(\boldsymbol{x},W) = d<W_{SRL}, W> = d(O_1,O_2) = |O_2-O_1|,
\end{equation}
where,  
$W$ refers to $W_h$ or $W_c$, calculated by Eq.~(\ref{Set2}) - (\ref{Set3}), respectively.
$\boldsymbol{x}$ is a 5-dimension decision vector.
The Euclid distance is used to describe the center position similarity. 
\end{itemize}
\begin{itemize}
    \item \textit{Semi Major-axis Distance (S-MAD): }
Based on manipulator ellipsoid \cite{MARIC2021103865,IJRR_Noemie}, we choose the semi major-axis vector of the ellipsoid to calculate the index. S-MAD means the similarity of the two ellipsoids, which is defined as,
\begin{equation}
\label{F2}
\begin{aligned}
\quad f_2(\boldsymbol{x},W) &= d<\overrightarrow{O_1 A_1},\overrightarrow{O_2 A_2}> \\
&= 1-{\rm cos\frac{\overrightarrow{O_1 A_1} \cdot \overrightarrow{O_2 A_2}}{|\overrightarrow{O_1 A_1}||\overrightarrow{O_2 A_2}|}},
\end{aligned}
\end{equation}
where $A_1, A_2$ are the points on the semi major axis. $\overrightarrow{O_1 A_1},\overrightarrow{O_2 A_2}$ are the semi major axis vectors of the SRL ellipsoid and the reference hand ellipsoid, respectively.
The cosine distance is used to describe the posture similarity. 
\end{itemize}

\begin{itemize}
    \item \textit{Oblateness (Obl): }
The oblateness is used to describe the shape similarity, which could be defined by
\begin{equation}
\label{F3}
\begin{aligned}
    Obl &= e_1e_2 = \frac{(a-b)(a-c)}{a^2},\\
\quad f_3(\boldsymbol{x},W) &= |{\rm ln}(\frac{Obl_2}{Obl_1})| = |{\rm ln}(\frac{e_{21}e_{22}}{e_{11}e_{12}})|,
\end{aligned}
\end{equation}
where, $e_1, e_2$ are the eccentricity multiplication of the ellipse projection of the ellipsoid along two major axes.
$a, b, c$ are the values of three semi major axis respectively.
The oblateness is used to describe the shape similarity. 
\end{itemize}
\begin{itemize}
    \item \textit{Volume (Vol): }
$vol_\mathcal{M}$ refers to the size of the ellipsoid, which is 
\begin{equation}
\label{F4}
\begin{aligned}
    vol & = \frac{4}{3}\pi abc,\\
    \quad f_4(\boldsymbol{x},W)&  = \frac{vol_{\mathcal Mref}}{vol_\mathcal{M}},
\end{aligned}
\end{equation}
where, $vol_\mathcal{M}$ is the volume of SRL workspace ellipsoid, $vol_{\mathcal Mref}$ is the volume of reference ellipsoid. The goal is to find an ellipsoidal space with a larger region volume.
The volume index is used to describe the size similarity. 
\end{itemize}

\subsubsection{Lower-SRL Static Force for STS ($F_{STS}$)}
The human body was composed of three parts: shin, thigh, and HAT (head, arm, and torso). 
It was assumed that the shin included the foot, and the toe joint was ignored.
Therefore, the human model was described as a triple-inverted pendulum \cite{HuoTRO2022}.

When used as the lower limb, the optimization objective is the robotic system output force $F_H$ actuated by motors, which can be calculated in supplenmentary material.
The computation of force $F_H$ takes into account both the length of the linkage and the stationary point on the ground relative to the person.
Then, it can be minimized by taking the reciprocal as the indicator, which is,
\begin{equation}
\label{F5}
f_5(\boldsymbol{x}) = \frac{\kappa}{F_{H}},
\end{equation}
where, $\kappa$ is a positive constant coefficient.

\subsubsection{Relative Mass (RM)}
In addition, indicators are required to define system mass and inertia. Since the total length of the system should not be less than the user's leg, there is a minimum length. A relative mass indicator is established to describe the impact of mass.
It is assumed that the material of the connecting rod is homogeneous, meaning the density is constant. The total mass index can be defined as follows,
\begin{equation}
\label{F7}
f_6(\boldsymbol {x})= |\sum^4_{i=1}\rho l_i-\rho L_0|,
\end{equation}
where, $l_i$ means the $i$-th linkage length, $\rho$ is the material density. $L_0$ denotes the overall length of the human leg and is used to ensure that the linkage length does not become shorter than the leg itself. 

\subsubsection{Moment of Inertia (MI)}
Since the mass of the connecting rod is much smaller than that of the joint and the end effector, the mass of the connecting rod is neglected.
We consider that the maximum moment of inertia relative to the first revolute $q_1$ to describe the inertia of the system (see Fig.~\ref{Kine}(b)), which is,
\begin{equation}
\label{F8}
f_7(\boldsymbol {x})= \sum^4_{i=1}m_{i+1}r_i^2,
\end{equation}
where, $m_{i+1}$ denotes the mass of $i+1$-th joint module or end effector, $r_i=(\sum^i_{j=1}l_j)^2$ is the vertical distance between $i+1$-th joint or end effector and axis of rotation.

\subsection{Solving Algorithm and Evaluation}
The characteristics of optimization problems outlined in this context are as follows: (1) The parameter is time-invariant, indicating a static optimization scenario. (2) The calculation of a single objective function is high dimension and non-linear, with multiple objective functions exhibiting non-convex behavior.
In this section, the proposed MSCFA is introduced in the first part. 
This approach enhances the global search capability of the algorithm, resulting in faster convergence.
The simulation results are compared with other SOTA solving methods in the second part, such as NSGA-II \cite{NSGA}, BOEA \cite{BOEA}, SMEA-PF \cite{SMEA-PF}, VMEF \cite{VMEF}, MSPSO \cite{MSPSO}, MEFA-CD \cite{MEFA-CD}.
\subsubsection{MSCFA Framework}
\begin{figure}[!t]
\centering
\includegraphics[width=1\columnwidth]{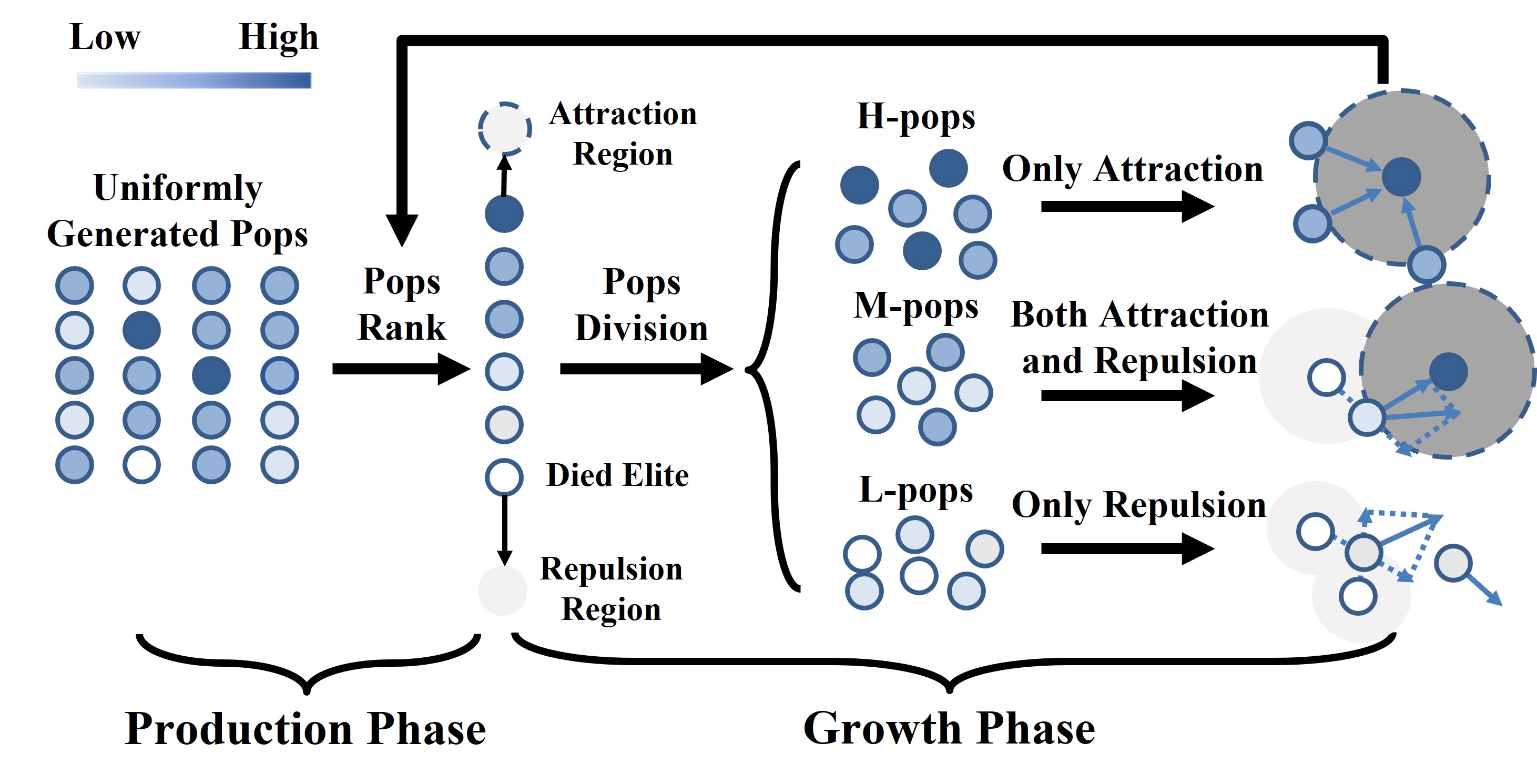}
\caption{Schematic diagram of proposed MSCFA: Each small circle represents an individual firefly in the algorithm. The darker the color of the circle, the higher the fitness of the firefly. Individuals with high fitness will generate an attraction region, represented by a dark-colored background circle with a dashed line. Conversely, individuals with low fitness will generate an repulsion region, represented by a light-colored background circle.}
\label{MSCFA}
\end{figure}

In the multi-subpopulation correction firefly algorithm, each firefly  represents a feasible solution, with its brightness indicating the solution's quality. Brighter fireflies attract dimmer ones, leading them toward better positions. 
As distance increases and brightness fades, the attraction correspondingly decreases, reducing the attraction range.
If a firefly is pareto dominated, it is removed, thereby creating a repulsion zone to prevent other fireflies from approaching.
Fig.~\ref{MSCFA} shows the schematic diagram of proposed MSCFA.
The specific process of the proposed MSCFA is presented in Algorithm \ref{MSFA}.
The production phase refers to the initialization process of the algorithm, and the growth phase denotes the iterative loop process. First, let us introduce some notations and concepts.

\textit{Definition 1, Pareto Dominate}: 
Given two vectors $\boldsymbol {x,y}\in \mathbb R^n$, $\boldsymbol x$ Pareto dominates $\boldsymbol y$ if satisfy $F(\boldsymbol x)<F(\boldsymbol y)$ and $f_i(\boldsymbol x)\leq f_i(\boldsymbol y)$ for $i= 1,2...m$, and there exist at least one index meets $f_i(\boldsymbol x)< f_i(\boldsymbol y)$.

if one firefly individual $\boldsymbol y$ is pareto dominated by another firefly individual $\boldsymbol x$, then the weaker firefly individual $\boldsymbol y$ would be eliminated and generates a new one near the firefly individual $\boldsymbol x$.
\begin{equation}
\label{pd}
\boldsymbol{y_i'} = \boldsymbol{x_i} + k(rand-0.5).
\end{equation}
\begin{itemize}
    \item 
\textit{Attraction region:} the basic firefly algorithm attraction region is given by
\begin{equation}
\label{fa1}
I(r) = I_0 e^{-\gamma r^2} + I_{min},
\end{equation}
where $I_0, I_{min}, \gamma$ are the system parameters to be set, $r$ refers to the distance between two individuals, which is defined as Cartesian distance,
\begin{equation}
\label{fa2}
r_{ij} = ||\boldsymbol{x_j-x_i}|| = \sqrt{\sum^d_{n=1}(x_{i,n}-x_{j,n})^2}.
\end{equation}
\end{itemize}
\begin{itemize}
    \item 
\textit{Repulsion region:}
firefly individual with pareto dominated will have a repulsion region to other fireflies, and the repulsion function is set to
\begin{equation}
\label{fa5}
\beta(r) = \beta_0 e^{-\eta r^2} + \beta_{min},
\end{equation}
where $\beta_0, \beta_{min}, \eta$ are the system parameters to be set.
\end{itemize}

\begin{algorithm}[t]
\caption{Multi-Subpopulation Correction Firefly Algorithm}
\label{MSFA}
\begin{algorithmic}[1] 
\renewcommand{\algorithmicrequire}{ \textbf{Input:}}
\Require 
$\boldsymbol {F(x)}$: the objective function; 
$\boldsymbol{x}$: a 5-D solution vector; 
$\Phi$: the MOO problem cost function 
\renewcommand{\algorithmicensure}{ \textbf{Output:}}
\Ensure 
$\boldsymbol {x_b}$: optimal individual;
$\Phi(\boldsymbol{x_b})$: optimal cost function
\State Initial population $P$ by Eq.~(\ref{IPG1}); 
\For{$Iter=$1 : MaxDT}
\State Calculate the workspace by Eq.~(\ref{Set1});
\State Calculate sub-objective function by Eq.~(\ref{F1})-Eq.~(\ref{F4}), Eq.~(\ref{F5}) and Eq.~(\ref{F7});
\State Calculate the fitness value by Eq.~(\ref{MOP2});
\State Rank the individuals according to correlation Eq.~(\ref{Div});
\State Check if there are pareto dominate relations, generate new fireflies by Eq.~(\ref{pd}) and repulsion region by Eq.~(\ref{fa5})
\State Population division into three sub-populations $SubP_H$, $SubP_M$ and $SubP_L$; 
\State Generate new $SubP_H$ by Eq.~(\ref{fa8}); 
\State Generate new $SubP_M$ by Eq.~(\ref{fa6}); 
\State Generate new $SubP_L$ by Eq.~(\ref{fa7}); 
\State $Iter$ = $Iter$ + 1;
\EndFor
\end{algorithmic} 
\end{algorithm}

\textit{a. Production Phase - Initial Population Generation Strategy:}
The initial population of firefly algorithm are mostly generated randomly, which cannot be well distributed in the search space. It will affect the speed of population convergence and the risk of falling into local optimum. 
The initial population resulting from a uniform distribution is as follows:
\begin{equation}
\label{IPG1}
\begin{aligned}
&\mathbf X_0 = \rm {meshgrid} (\boldsymbol{x_1},\boldsymbol{x_2},\boldsymbol{x_3},\boldsymbol{x_4},\boldsymbol{x_5}),\\
&\boldsymbol x_i \sim {\rm U}(\mathbf{LB}, \mathbf{UB}), i=1,2,3,4,5.
\end{aligned}
\end{equation}

\textit{b. Growth Phase - Population Division and Update:}
The proposed algorithm divides the population into three sub-populations with different fitness values by utilizing correlation information.
\begin{equation}
\label{Div}
\psi = ||\Phi(\boldsymbol{x_i}) - \Phi(\boldsymbol{x_b})||,
\end{equation}
where, $||\cdot||$ is the Euclidean distance, $x_b$ is the best individual of the population.
\begin{itemize}
    \item 
\textit{High correlation sub-population:}
The update law of high correlation sub-population $SubP_H$ is 
\begin{equation}
\label{fa8}
\boldsymbol{x_{i+1}} =\boldsymbol{x_i}+I(r_{ij})(\boldsymbol{x_j}-\boldsymbol{x_i})+\epsilon_i.
\end{equation}
\end{itemize}
\begin{itemize}
    \item 
\textit{Medium correlation sub-population:}
The update law of medium correlation sub-population $SubP_M$ is 
\begin{equation}
\label{fa6}
\begin{aligned}
\boldsymbol{x_{i+1}} &=\boldsymbol{x_i}+I(r_{ij})(\boldsymbol{x_j}-\boldsymbol{x_i})-\beta(r_{ik})(\boldsymbol{x_k}-\boldsymbol{x_i})+\epsilon_i,\\
&= (1-I(r_{ij})+\beta(r_{ik}))\boldsymbol{x_i} + I(r_{ij})\boldsymbol{x_j}-\beta(r_{ik})\boldsymbol{x_k}\\
&\quad +\epsilon_i.
\end{aligned}
\end{equation}
\end{itemize}
\begin{itemize}
    \item 
\textit{Low correlation sub-population:}
The update law of low correlation sub-population $SubP_L$ is 
\begin{equation}
\label{fa7}
\boldsymbol{x_{i+1}} = \boldsymbol{x_i}-\beta(r_{ik})(\boldsymbol{x_k}-\boldsymbol{x_i})+\epsilon_i.\\
\end{equation}
\end{itemize}

\subsubsection{Algorithm Evaluation}
In this section, simulation results are provided to illustrate the superiority of the proposed MSCFA algorithm over other basic methods.
The model parameters were set to : $num(W) = 10000, {\rm Pop} = 81, {\rm MaxDT} = 300, \alpha = 0.5, I_0 = 1, I_{min} = 0.2, \gamma = 1, \beta_0 = 1, \beta_{min} = 0.01, \eta = 10$.
\begin{figure}[!t]
\centering
\includegraphics[width=1\columnwidth]{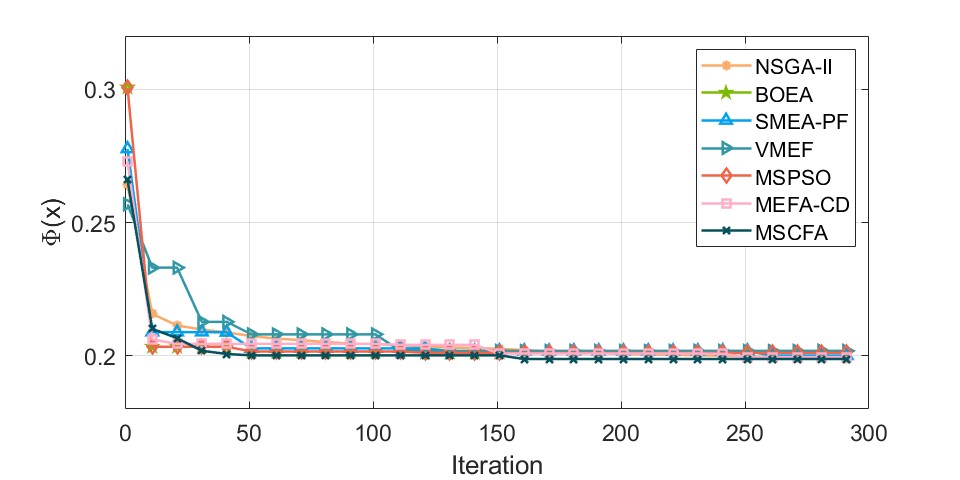}
\caption{Converge comparison between different SOTA solving method results.}
\label{comparison}
\end{figure}

\begin{figure}[!t]
\centering
\includegraphics[width=1\columnwidth]{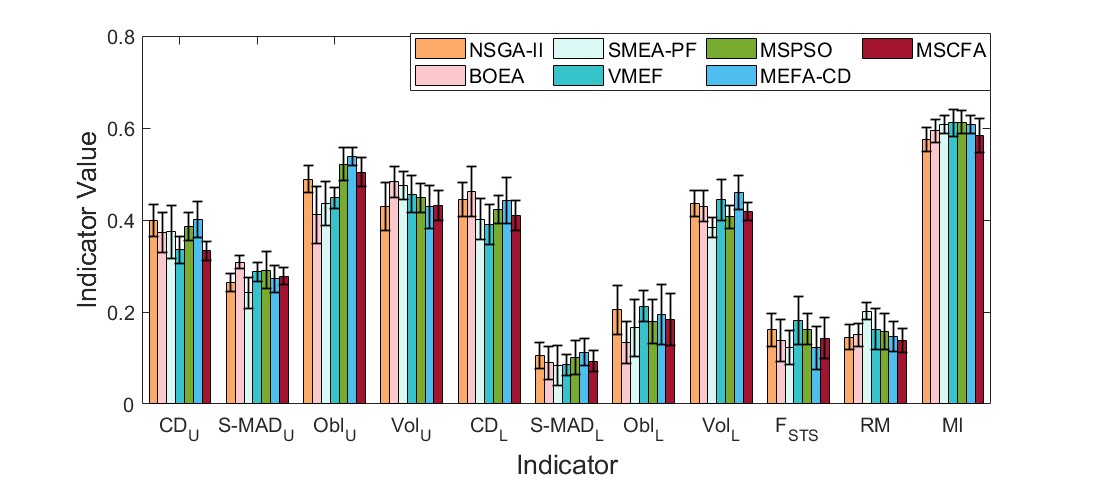}
\caption{Illustration of proposed indicators values. Upper limb workspace indicators: $\rm CD_U, S-MAD_U, Obl_U, Vol_U$. Lower limb workspace indicators: $\rm CD_L, S-MAD_L, Obl_L, Vol_L$. Static indicator: $\rm F_{STS}$ and system property indicators: $\rm RM,MI$.}
\label{chart}
\end{figure}

\begin{table*}[!t]
\caption{Performance Comparison among Different SOTA Solving Algorithms}
\label{Tab1}
\centering
\renewcommand\arraystretch{1.4}
\begin{threeparttable}
\begin{tabular}{c c c c c c c c}
\hline
\hline
Algorithm & MaxDT$^*$ & Pop$^*$  & Best Cost & Best Solution  & Total Length & Conv Iter$^*$  & Time Cost (per iter) \\
\hline
NSGA-II \cite{NSGA}  & 300 & 81 & $0.1994 \pm 0.004$ & [0.100, 0.401, 0.305, 0.209, 0.200] & 1.015 m & $180 \pm 30$ & $1948.2 \pm 26.2$ s\\
BOEA \cite{BOEA} & 300 & 81 & $0.2008 \pm 0.009$ & [0.100, 0.413, 0.291, 0.200, 0.226] & 1.004 m & $152\pm 27$ & $988.4 \pm 27.8$ s\\
SMEA-PF \cite{SMEA-PF} & 300 & 81 & $0.2002 \pm 0.008$ & [0.100, 0.405, 0.311, 0.190, 0.212] & 1.006 m & $218\pm 43$ & $1044.9 \pm 56.4$ s  \\
VMEF \cite{VMEF} & 300 & 81 & $0.2018 \pm 0.009$ & [0.100, 0.424, 0.317, 0.189, 0.184] & 1.030 m & $132\pm 23$ & $833.8 \pm 24.3$ s \\
MSPSO \cite{MSPSO} & 300 & 81 & $0.2011 \pm 0.015$ & [0.100, 0.417, 0.309, 0.214, 0.220] & 1.040 m & $147\pm 36$ & $922.5\pm 19.9$ s \\
MEFA-CD \cite{MEFA-CD} & 300 &  81 & $0.1995 \pm 0.015$ & [0.100, 0.394, 0.317, 0.210, 0.216] & 1.021 m & $261 \pm 16$ &$ 683.0 \pm 15.4$ s  \\
MSCFA & 300 &  81 & $\mathbf{0.1993 \pm 0.012}$ & [0.100, 0.403, 0.296, 0.204, 0.198] & $\mathbf{1.003\ m}$ & $\mathbf{130 \pm 25}$ & $\mathbf{634.2\pm 16.7\ s}$  \\
\hline
\hline
\end{tabular}
\begin{tablenotes}
\item[*] MaxDT: Maximum Iteration; Pop: Populations; Conv Iter: Convergence Iteration
\end{tablenotes}
\end{threeparttable}
\end{table*}

The simulation environment is configured by Matlab R2023a. 
Each algorithm was executed 10 times and the results were averaged.
Tab.~\ref{Tab1} displays the performance of solving solutions across several algorithms.
The proposed algorithm converges to around $[0.1, 0.4, 0.3, 0.2, 0.2]^T$ from a solution distribution standpoint, showing minimal practical variation. 
Fig.~\ref{comparison} compares common solving methods with the proposed MSCFA method based on the results.
The convergence iteration index is based on the number of iteration during which the loss function value remains stably below 0.2.
On average, the suggested strategy reaches the ideal value around the 130th generation and is the quickest in terms of time cost index.
Fig.~\ref{chart} displays the average results from repeated trials
with the same initial conditions. The indicators are defined in section II-B.
Although the proposed approach may not be optimal for each individual item, it still achieves an overall optimal result, indicating its superiority in providing a comprehensive solution.   
It is observed that the proposed method achieves the shortest running time while converging to the same level of accuracy, demonstrating its superiority in convergence speed.

\section{Simulations Optimization Results}
\begin{figure}[!t]
\centering
\includegraphics[width=0.5\textwidth]{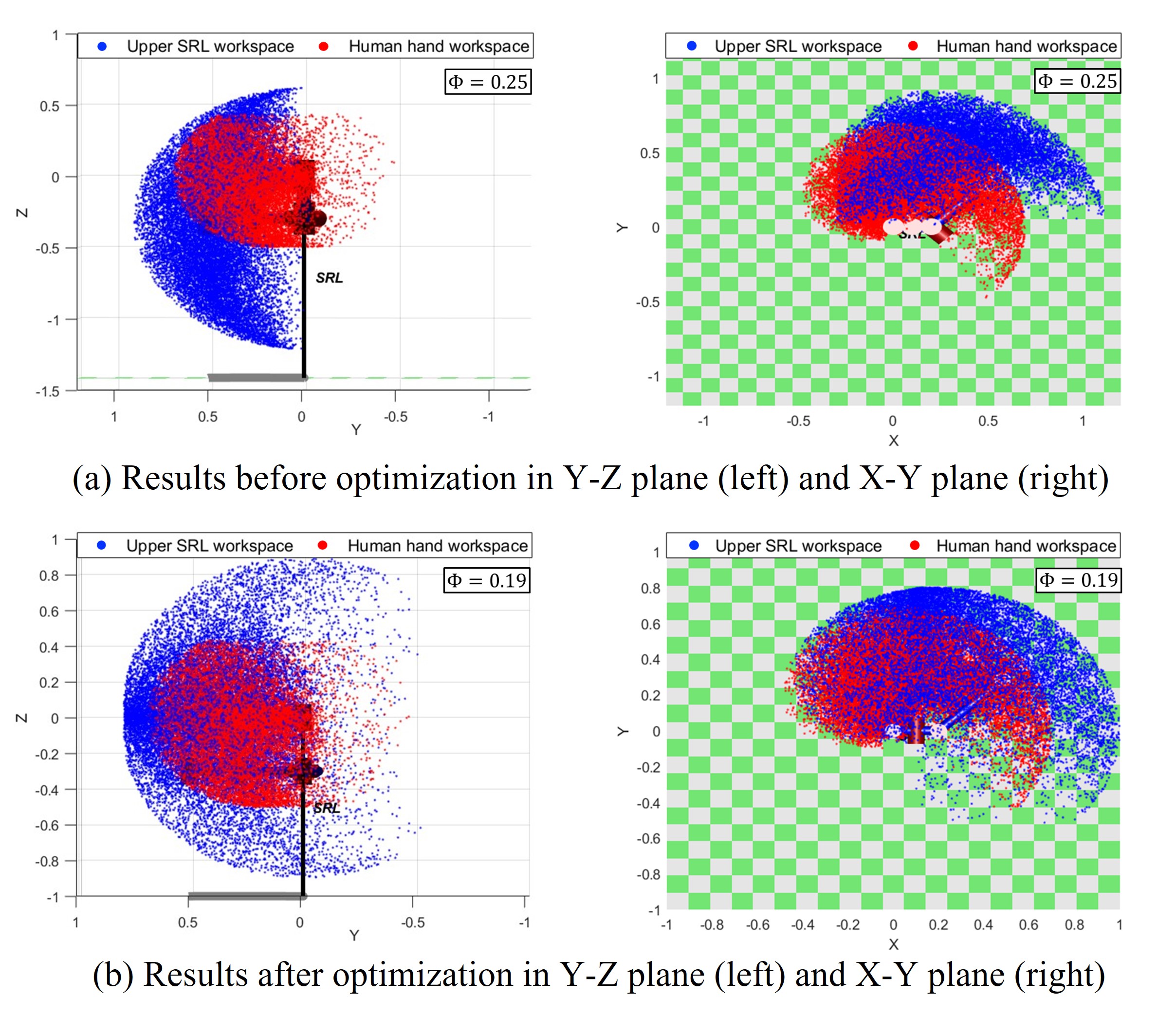}%
\caption{Comparison of the upper SRL workspace and human upper limb workspace results before and after optimization from different perspectives. The blue points represent the workspace of the SRL, while the red points correspond to the human upper limb. }
\label{Re1}
\end{figure}

\begin{figure}[!t]
\centering
\includegraphics[width=0.5\textwidth]{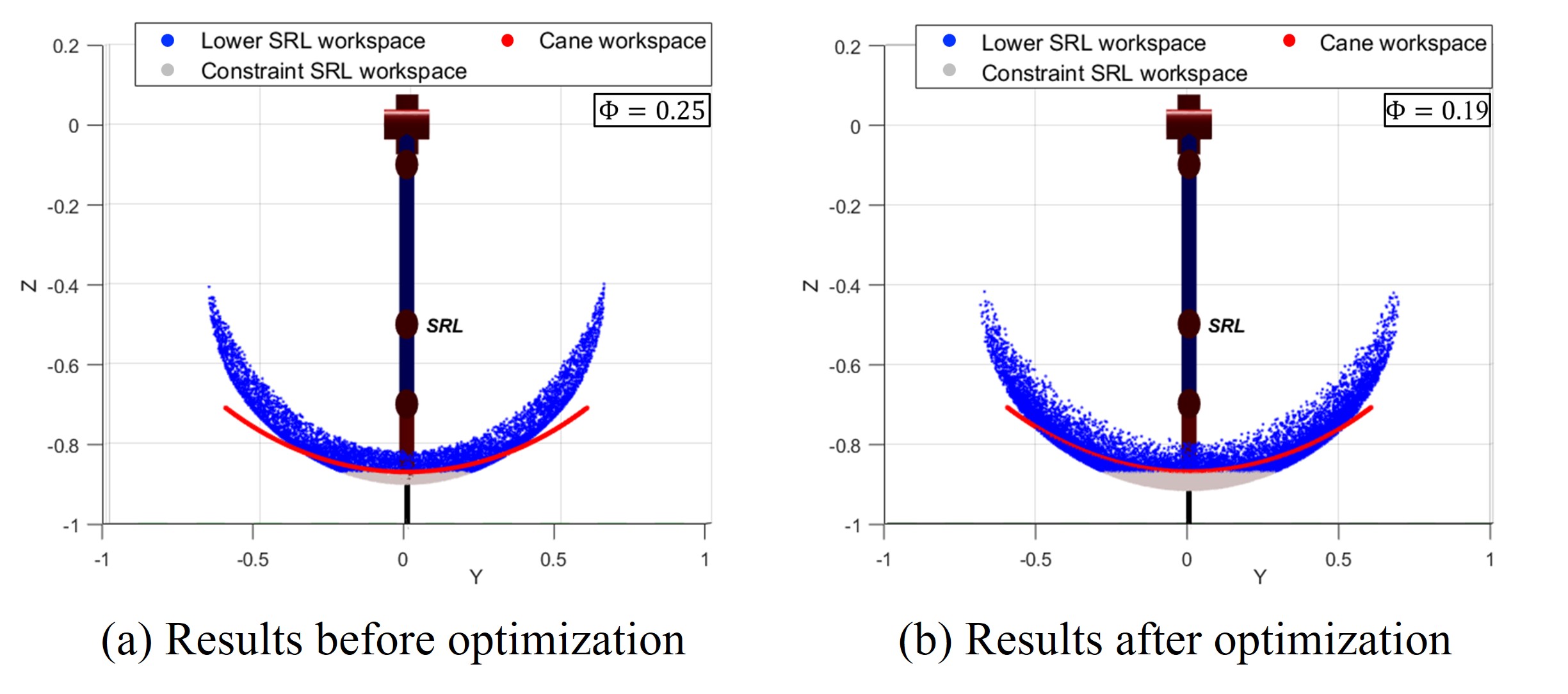}%
\caption{Comparison of the lower SRL reduced workspace and cane assisted workspace results before and after optimization. The blue points represent the workspace of the SRL, and the gray points are the unreachable workspace due to constraints. The red arc refers to the cane.}
\label{Re2}
\end{figure}

\begin{figure}[!t]
\centering
\includegraphics[width=0.4\textwidth]{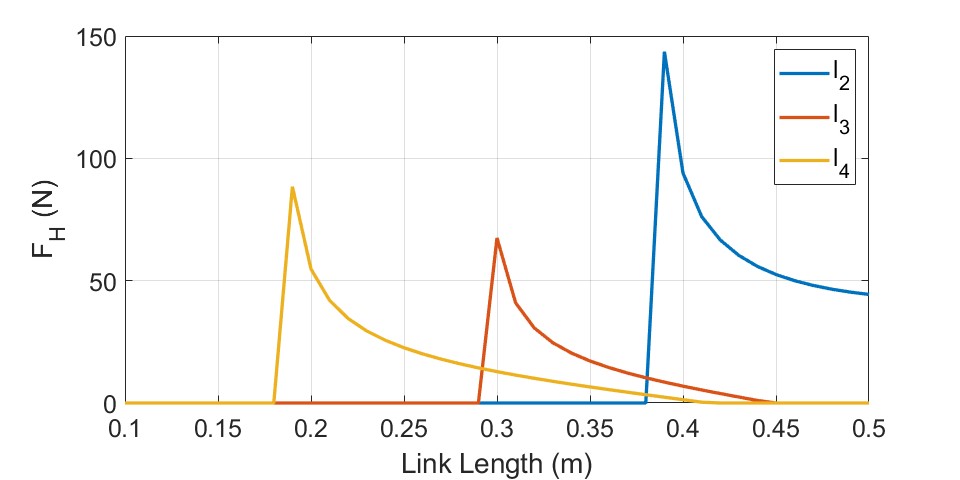}%
\caption{The variation trend diagram of maximum STS static force $F_H$ with different connecting rod lengths, while other links remain constant.}
\label{Re3}
\end{figure}
In the simulation, we set the total length of the human leg as 0.9 m, denoted as $L_0 = 0.9$ m.
Based on the results in section II-C, the link lengths are configured as $[0.10, 0.40, 0.29, 0.21]^T$, with $c=0.19$ m. 
The rest of the simulation parameters are set as shown in section II-C, part 2.
Subsequently, simulation calculations were conducted using Matlab R2023a to obtain the results related to the workspace of the SRL.
As a comparison with the optimal parameters, the mechanism parameter is configured as $[0.25, 0.25, 0.25, 0.25]^T$, with $c=0$ m.

Fig.~\ref{Re1} and Fig.~\ref{Re2} show the workspace results from different perspectives. 
Each workspace point is determined by assigning a random set of joint angles using the Monte Carlo method.
The figures verify that a set of workspace points in a crescent shape can be enveloped by an ellipsoid \cite{Kensuke,WU2020103711}.
Fig.~\ref{Re1} shows the comparison of the upper SRL workspace and human upper limb workspace results before and after optimization from different perspectives.
It can be seen that, compared to the results before optimization, the optimized SRL workspace is closer to the actual workspace of the human upper limb. This is reflected in the quantitative indicator $\Phi$ (defined in Eq.~(\ref{MOP2})), which decreases from 0.25 to 0.19.
Fig.~\ref{Re2} shows the comparison of the lower SRL workspace and cane assisted workspace results before and after optimization in Y-Z plane.
The optimized SRL workspace can better cover the required workspace points around the edges.
Fig.~\ref{Re3} shows the maximum STS static force result when only one link length is varied.
It can be seen that the optimal value of rod length is distributed around $l_2=0.4$ m, $l_3 = 0.3$ m, $l_4 = 0.2$ m, which is consistent with the solution result of the MSCFA algorithm.
The results indicate that the proposed method has outstanding problem-solving capabilities and provides a solid theoretical foundation for mechanism design.

\section{Experiments and Discussions}
\label{Sec_Mech}
In this section, the mechanical design of the general-purpose SRL is introduced. Detailed simulations and experiments are presented to evaluate the proposed optimization strategy across limb functions. 
Additionally, validation experiments are conducted for the proposed general-purpose SRL mechanism. 

\begin{figure}[!t]
\centering
\includegraphics[width=0.9\columnwidth]{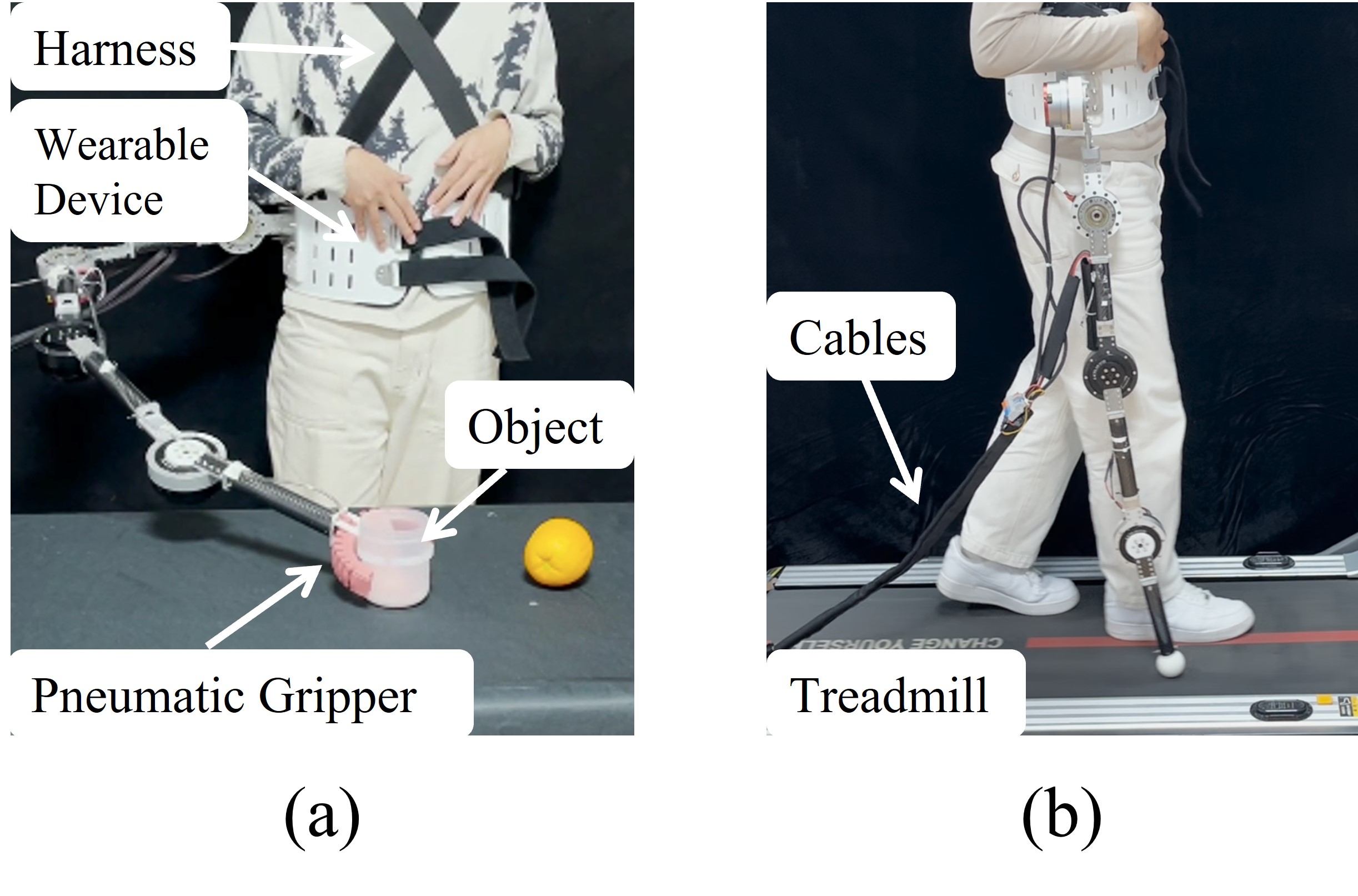}
\caption{Prototype setup of SRL. (a) The general-purpose SRL using as upper limb function. (b) The general-purpose SRL using as lower limb function.}
\label{Prototype}
\end{figure}

\begin{table}[!t]
\caption{Hardware Specifications}
\label{Tab3}
\centering
\renewcommand\arraystretch{1.3}
\tabcolsep=0.03\columnwidth
\begin{tabular}{c c c}
\hline
\hline
Type & Parameter & Value\\
\hline
TMotor & Weight & 521 g \\
        & Size & $\Phi 83\times 50$ \\
        & Rated torque & 8.3 Nm\\
        & Peak torque & 24.8 Nm\\
        & Rated speed & 310 rpm\\
MintaSCA & Weight & 1 kg\\
        & Size & $\Phi 80\times 65.7$ \\
        & Rated torque & 38 Nm\\
        & Peak torque & 108 Nm\\
        & Rated speed & 29.7 rpm\\
Links   & Lengths & [0.1 m,  0.4 m,  0.3 m,  0.2 m]\\
        & Material & Carbon fiber pipe \\
        & Material density & 1.65 $\rm g/cm^3$ \\
        & Mass &  [20.8 g, 83.3 g, 62.5 g, 41.6 g] \\
System  & Mass & 5.0 kg\\
\hline
\hline
\end{tabular}
\end{table}

\subsection{Mechanical Design of General-purpose SRL}
The prototype includes four motors, as depicted in Fig.~\ref{Prototype}. 
Proximal joints require higher torque output, whereas end joints of the manipulator need to respond rapidly to ensure flexible movement.
There are two types of robotic joint module motors. The first one (TMotor AK70-10, CubeMars) is a high-speed motor with low output torque. The second one (QDD Pro-PR60-100-80, MintaSCA) is a low-speed motor with high output torque. 
The hardware specifications are displayed as Tab.~\ref{Tab3}.
Each module motor is equipped with a position encoder to provide feedback rotating angle.
Carbon fiber tubes are used as the primary material for connecting the joints.
The motor is compatible with CAN communication mode. The signal acquisition board is the DAQ-USB-6001 (National Instrument Ltd.).
The total height between the harness and the ground in lower limb function mode is about 1.0 m. The total weight of the system is 5.0 kg.

The pneumatic flexible gripper at the end is constructed from silicone rubber. When the gas is pumped in, the rubber material bends to grasp the object. When the air is deflated, the gripper releases the object \cite{ru2023JoVE}. The air pressure of the filling gas is set at approximately 150 kPa.    

\begin{table}[!t]
\caption{Details of Participating Healthy Participants}
\label{Subjects}
\centering
\renewcommand\arraystretch{1.3}
\tabcolsep=0.04\columnwidth
\begin{tabular}{c c c c c}
\hline
\hline
Subjects & Gender & Age & Height (cm) & Weight (kg)\\
\hline
S1 & Male & 29 & 178 & 70 \\
S2 & Male & 28 & 175 & 70 \\
S3 & Male & 23 & 177 & 75 \\
S4 & Male & 25 & 176 & 60 \\
S5 & Male & 25 & 175 & 70 \\
S6 & Male & 28 & 180 & 75 \\
\hline
\hline
\end{tabular}
\end{table}

\begin{figure}[!t]
\centering
\includegraphics[width=0.88\columnwidth]{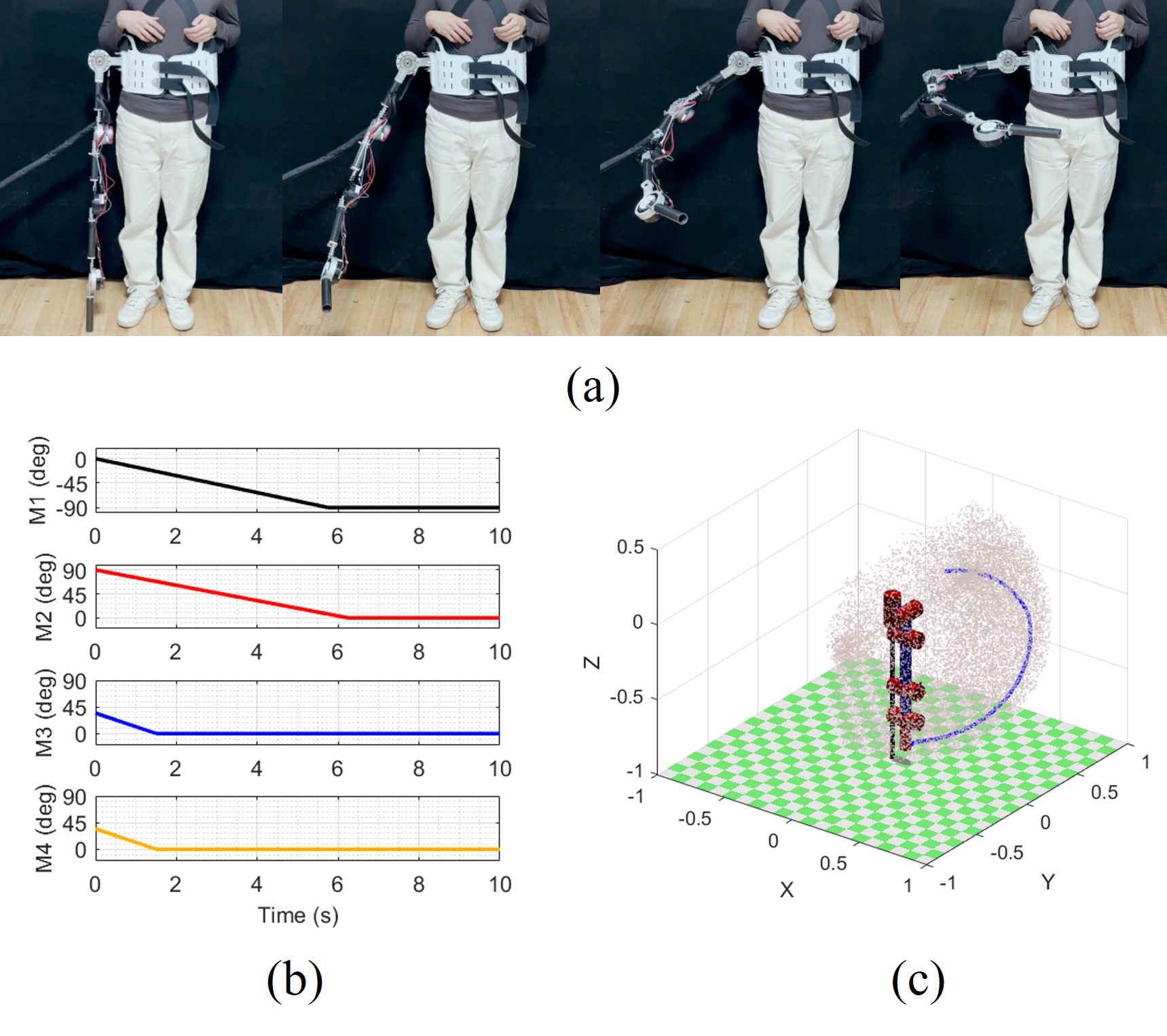}%
\caption{Task 1: Upper SRL convert to lower SRL. 
(a) Postures sequences demonstrated in UTL. 
(b) Joint rotation angle of SRL mechanism. 
(c) The workspace (grey) and the trajectory of manipulator end (blue).}
\label{Task1}
\end{figure}

\begin{figure*}[!t]
\centering
\includegraphics[width=0.88\textwidth]{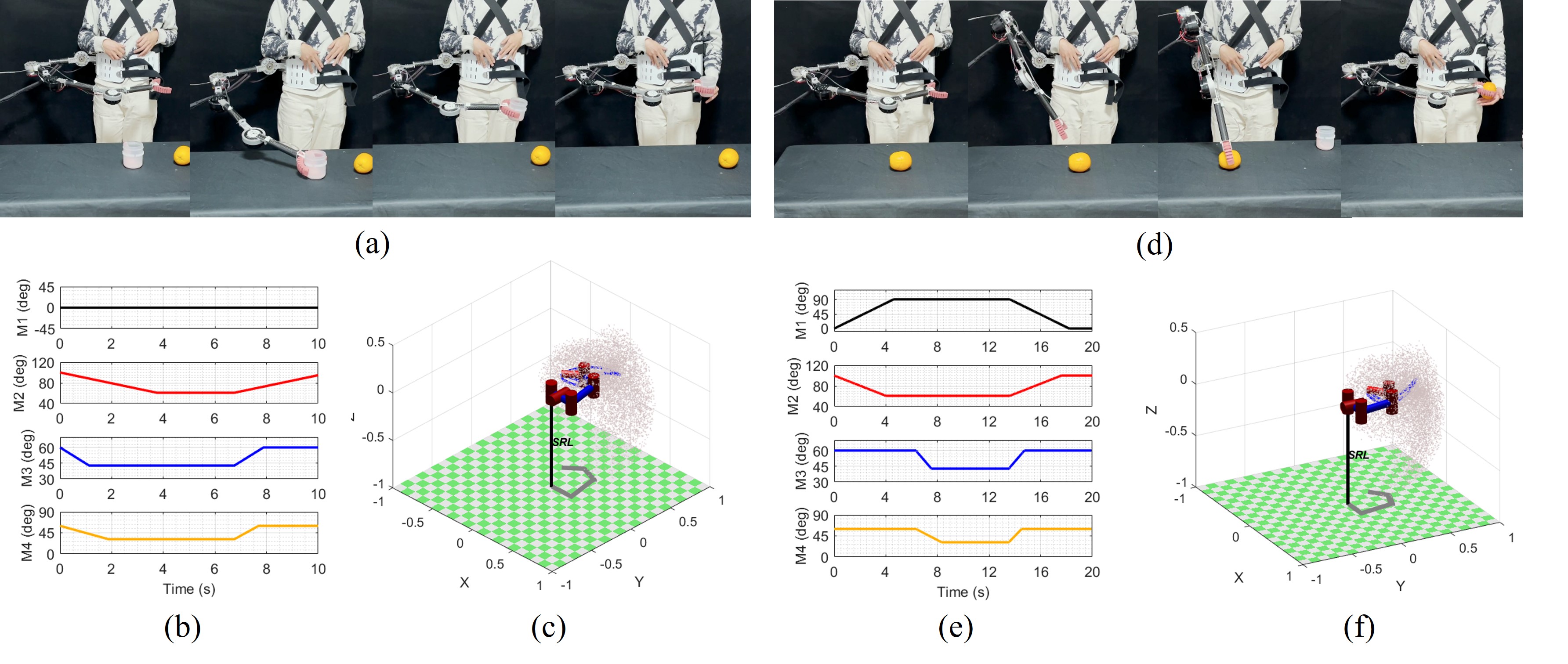}%
\caption{Task 2: Grasp an object under horizontal posture. 
(a) Postures sequences demonstrated in horizontal grasp. 
(b) Joint rotation angle of SRL mechanism. 
(c) The workspace (grey) and the trajectory of manipulator end (blue).
Task 2: Grasp an object under vertical posture.
(d) Postures sequences demonstrated in STS. 
(e) Joint rotation angle of SRL mechanism. 
(f) The workspace (grey) and the trajectory of manipulator end (blue).}
\label{Task2}
\end{figure*}

\begin{figure}[!t]
\centering
\includegraphics[width=1\columnwidth]{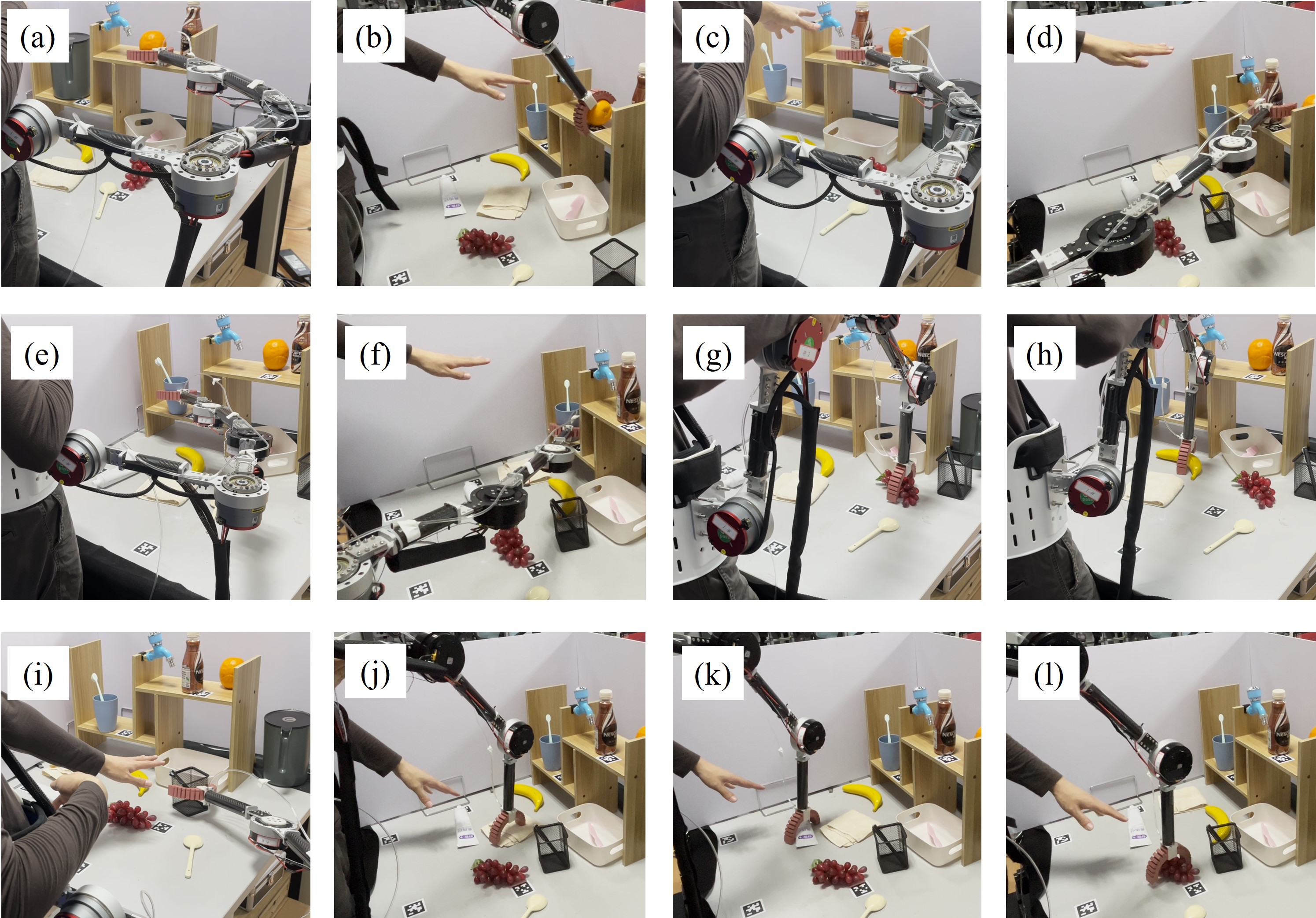}
\caption{The study examines the gripping capability of numerous objects in various postures of the general-purpose SRL, when used as the upper limb function.}
\label{AGO}
\end{figure}

\subsection{Implementation}
In our experiments, six healthy subjects were enrolled, as detailed in  Tab.~\ref{Subjects}.
A wide cup and an orange were used as the objects to be grasped (Fig.~\ref{Prototype}(a) and Fig.~\ref{Task2}(a)(d)).
The subject wore the general-purpose SRL while walking on a treadmill (Fig.~\ref{Prototype}(b) and Fig.~\ref{Task3}(a)(b)). The walking speed was set to 2.0 km/h.
We collected joint space trajectories and developed offline reference trajectory for the SRL mechanism.

The ethics approval for experiments with subjects was granted by the Ethics Committee of Tongji Medical College, Huazhong University of Science and Technology (NO. IORG0003571).

To ensure safety and reliability, the SRL was first tested without subjects prior to the formal experimental program. The following amendments were incorporated into the approved protocol, which had been reviewed and approved by No. IORG0003571.

1) Ensure that the system is set up in an isolated room with adequate power supply and minimal interference from other daily activities.

2) Participants are expected to have basic knowledge and experience in interacting with wearable robots. Personnel are required to perform strap adjustments, device operation, and demonstrate proficiency in the SRL robot's functionalities. Additionally, they must read and understand the instructional manual for the fitting program, follow instructions to adjust the fit with the robot, and know how to shut down the system in case of an emergency.

\begin{figure}[!t]
\centering
\includegraphics[width=1\columnwidth]{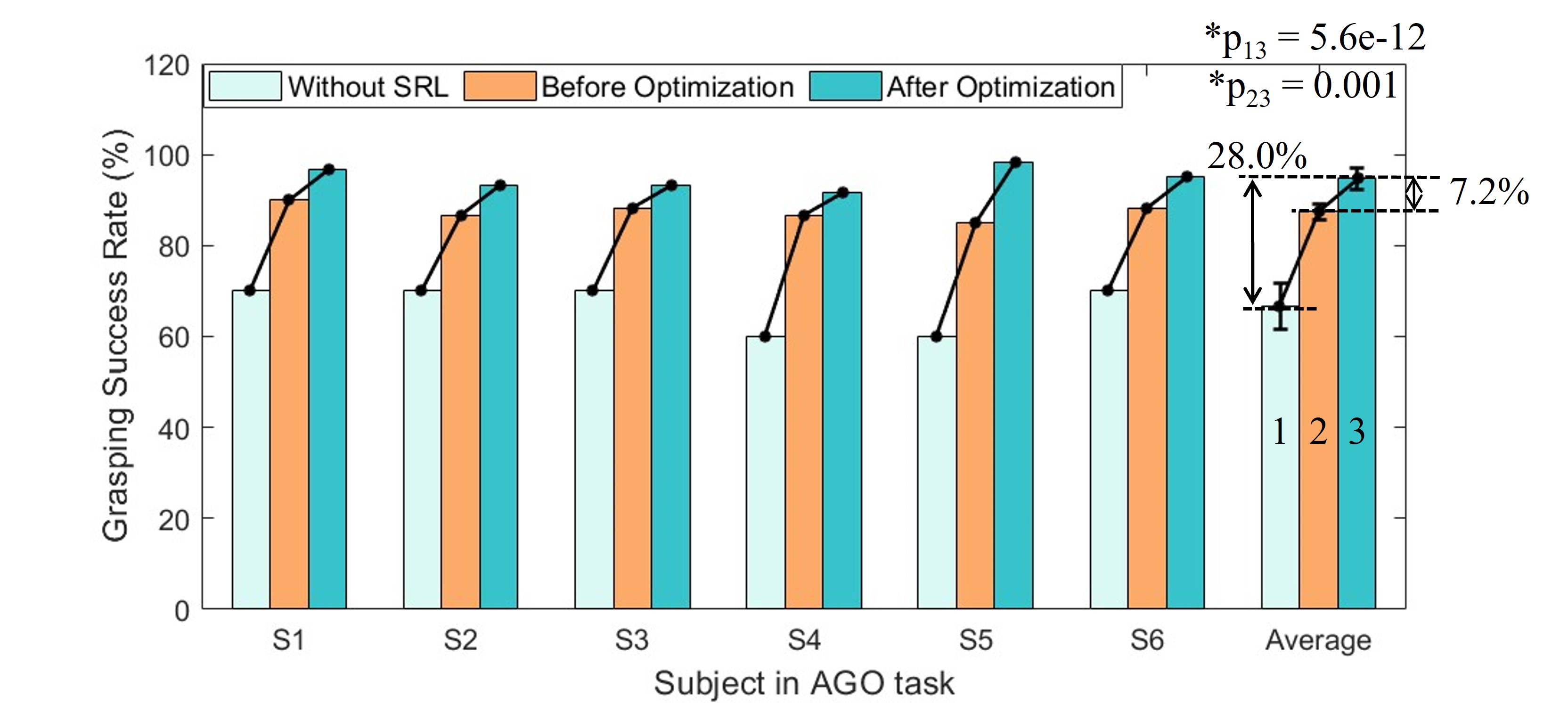}%
\caption{Comparison of grasp performance in AGO Task. The subscripts of the p-value represent the significance comparison of the two columns in the average group. The numbered label is shown in the rightmost column of the figure. For example, $\rm p_{13}<0.05$ means there is a significant difference between the first group (Without SRL) and the third group (After Optimization).}
\label{AGO1}
\end{figure}
The experiment procedures are as follows: The experiment begins with the experiment staff assisting the subject in wearing the SRL device and adjusting the straps for a comfortable fit. Next, the pneumatic gripper is installed at the end of the mechanism, and the subject performs the grasping task. 
The subject is instructed to stand in place, and sequentially grasp objects from the table, and place each object into his hand afterward. The completion of this sequence is recorded as one grasping task.
Following this, the experiment staff removes the pneumatic gripper and installs the lower limb end mechanism. The treadmill is then started and set to a speed of 2.0 kph, allowing the subject to proceed with the walking task. 
The subject is required to walk on the treadmill at a constant speed for 3 minutes, during which the SRL device follows each leg once.
Upon completion, the treadmill is turned off. Finally, the subject performs the STS task.
The subject is required to sit on a chair and slowly rise to a standing position, completing the STS motion. This process is repeated 10 times. 
After completing all tasks, the subject unfasten the straps and remove the device.

For each condition, we calculated the mean and standard deviation (SD). One-way ANOVA tests (significance level $\alpha$ = 0.05) was conducted to evaluate differences across conditions. These analyses were performed using MATLAB. Relevant methodological details and results are now presented in the corresponding experimental sections.

\begin{figure*}[!t]
\centering
\includegraphics[width=0.9\textwidth]{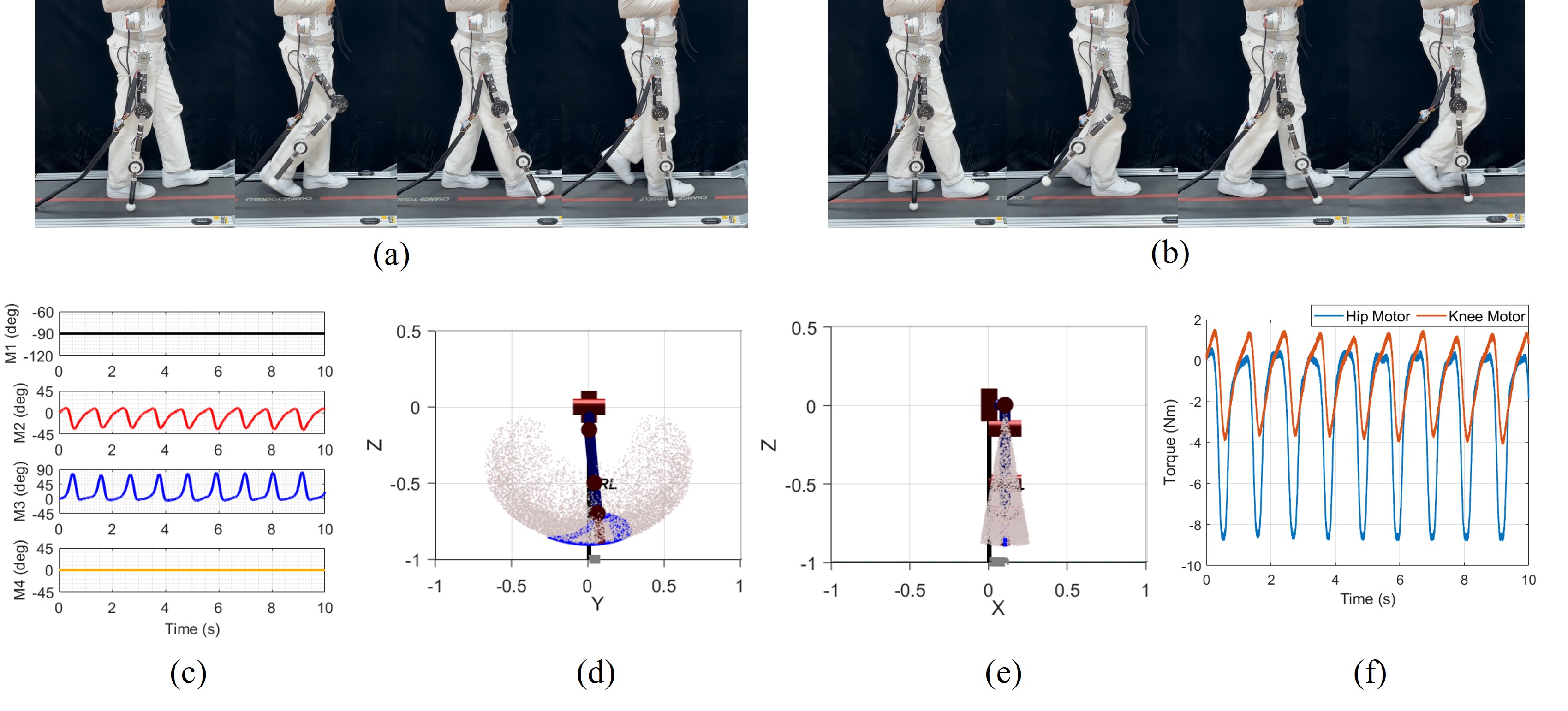}%
\caption{Task 3: WFL assistance.  
(a) Postures sequences demonstrated in following the right leg. 
(b) Postures sequences demonstrated in following the left leg.
(c) Joint rotation angle of SRL mechanism. 
(d) The workspace (grey) and the trajectory of manipulator end (blue) in side view.
(e) The workspace (grey) and the trajectory of manipulator end (blue) in rear view.
(f) The dynamic assistant torque.}
\label{Task3}
\end{figure*}

\begin{figure*}[!t]
\centering
\includegraphics[width=0.9\textwidth]{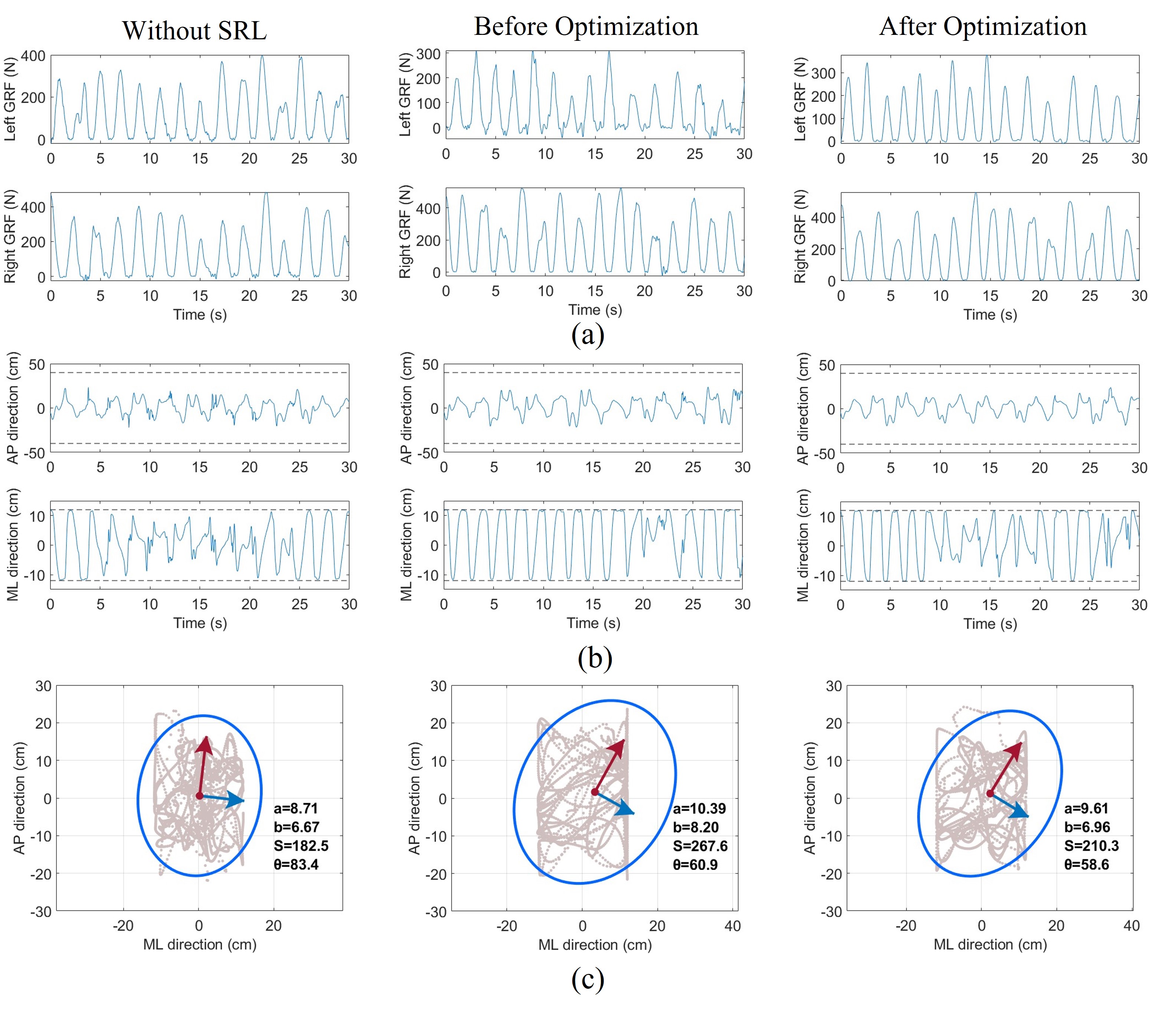}%
\caption{Comparison of without SRL support, before optimization and after optimization in WFL task. (a) GRF of double legs. (b) CoP in ML and AP direction. (c) CoP Illustration of the 95\% confidence ellipse.}
\label{CoP}
\end{figure*}

\subsection{Experimental Results}
Four tasks, including functional switch between upper and lower limb (UTL), assisting in grasping objects (AGO), walking following one leg (WFL), and STS movement were tested in experiments involving human participants.

\subsubsection{UTL functional switch}
Fig.~\ref{Task1}(a) shows the series of postures demonstrated during the UTL process experiments.
Once the transition between upper and lower limb functions begins, the initial joint rotation near the stationary end of the mechanism is mostly responsible for carrying out the planned movement. 
For reducing the bearing torque at the fixed end, it is recommended to prioritize rotating the third and fourth joints at the end. This helps achieve the desired reduction in torque impact (refer to Fig.~\ref{Task1}(b), where M3 and M4 cease to act at 1.5 s). 
The first and second joints rotate until they reach a stable position in the upper limb posture (Fig.~\ref{Task1}(b), where the movement stops around 6 s). The trajectory of manipulator end is shown in Fig.~\ref{Task1}(c).

During the movement of the mechanism, there is no interference with the natural workspace of the human body.
The planning results are precisely adapted to fit the natural human movement and operational task space.

\subsubsection{AGO movement}
Fig.~\ref{Task2}(a)(d) shows the series of postures conducted during the grasping mode experiments. 
Two grasping modes are provided. One is the grasping mode of horizontal posture (Fig.~\ref{Task2}(a)), and the other is the grasping mode of vertical posture (Fig.~\ref{Task2}(d)).

During the horizontal posture mode, the first joint of the proximal fixed end is locked. The two end joints move initially to decrease the torque (referring to Fig.~\ref{Task2}(b), M3 and M4 stop the action at approximately 2 s), then joint 2 moves and adjusts to the predetermined position, and finally, the pneumatic actuator completes the gripping operation.
The trajectory of manipulator end is shown in Fig.~\ref{Task2}(c).

When using the vertical posture grasp mode, the mechanism needs to first be adjusted to the vertical state (Fig.~\ref{Task2}(d)), and the subsequent operations are similar to those of the horizontal posture grasp. Fig.~\ref{Task2}(e) shows the joint rotation angle movement during the grasping process.
The trajectory of manipulator end is shown as Fig.~\ref{Task2}(f).

The planned trajectory fulfills the criteria of the SRL dexterous workspace in both grasping modes and effectively accomplishes the assigned task. The planning outcome aligns with the natural movements of the human body and the operational task space.
Fig.~\ref{AGO} demonstrates that the designed SRL mechanism, when employed as an upper-limb extension, not only retains grasping functionality but also expands the workspace beyond natural human capabilities.
The grasping success rate is used as a key metric to evaluate the system’s performance in grasping tasks.
A higher success rate indicates that the given configuration is easier to operate. In the experiment, each participant was required to perform six grasping tasks for each of total 10 objects, resulting in a total of 60 grasping tasks per participant. 
The task order is randomized to minimize the effects of familiarity with the grasping positions.
In this experiment, since the SRL is mounted on the waist of a subject, any lower-body movement may affect the upper limbs and the SRL.  Participants are instructed to keep their lower body in a fixed position to ensure a fair comparison. 
Participants are also allowed to grasp objects using various natural postures, including upper body bending, to extend their reach. This demonstrates that, under natural conditions, regardless of how the human body moves, the SRL can effectively expand the human limb’s workspace to reach areas that are inaccessible to human hands.
Our objective function aims to maximize the SRL workspace while ensuring it encompasses the natural upper limb workspace.
Consequently, the proposed SRL can serve as a substitute for the human limb and has the potential to replace the affected limb and hand of hemiplegia patients.

Fig.~\ref{AGO1} presents the statistics of six participants who completed the grasping task without and with SRL assistance, both before and after optimization. 
The success rate of the optimized system is significantly higher than that of the pre-optimized version, with an average increase of 7.2\% ($\rm p_{23}$ = 0.001).
This improvement is attributed to the larger workspace achieved after optimization, which provides a greater grasping range. Even with occasional grasping failures, the optimized system demonstrates superior grasping capabilities overall.

\subsubsection{WFL assistance}
Fig.~\ref{Task3}(a)(b) shows the series of postures during the walking experiments. Since the only difference between following the left and right legs is the phase, we merged the two into  one single task.

During lower limb walking, the proximal joint nearest to the fixed end remains locked. Joint 2 and joint 3 correspond to the hip joint and knee joint of the human leg, respectively. The preset motion trajectory of these joints is determined by analyzing and processing the human joint trajectory (Fig.~\ref{Task3}(c) and (d)). The fourth joint is secured in a locked position to maintain the output torque.
The trajectory of manipulator end is shown as Fig.~\ref{Task3}(d)(e). 
Fig.~\ref{Task3}(f) shows the estimated dynamic assistant torque during walking. Refer to supplenmentary material for specific calculation methods.  

To evaluate the performance of the system in WFL task, the Center of Pressure (CoP) metric was applied to reflect the dynamic balance and stability of the human-SRL system.
Surface electromyography (sEMG) was used to evaluate the biomechanical impact of the SRL on the subject. 
In the WFL task, muscle activation was measured in the calf's soleus (SOL) and medial gastrocnemius (MG). 
Maximum voluntary contractions (MVC) are obtained by asking participants to perform isometric contractions against manual resistance for each target muscle, with the highest 1-second moving average used for normalization.
During the experiment, participants were instructed to walk on a treadmill at a speed of 2.0 km/h for 3 minutes. Data from the stable 2-minute segment were used for statistical analysis. CoP positions and muscle activity in the calf were measured simultaneously. To quantify CoP stability, the 95\% confidence ellipse method was applied to calculate the CoP ellipse area and CoP sway direction \cite{schubert2014ellipse}. The average muscle activity during the motion was characterized using the Mean Amplitude Value (MAV), a commonly applied metric in biomechanics analysis \cite{merletti1999standards}.
\begin{figure}[!t]
\centering
\includegraphics[width=1\columnwidth]{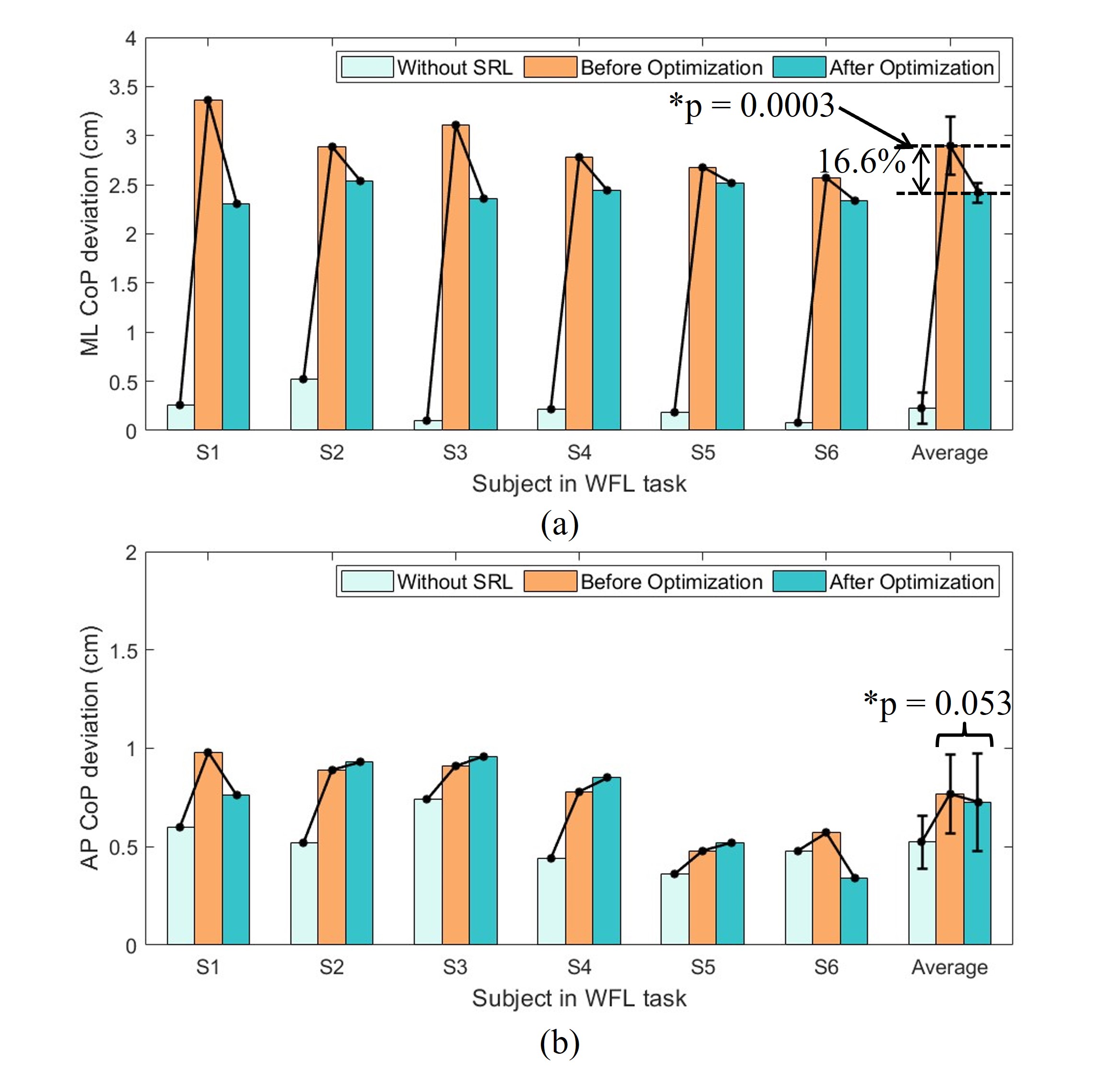}%
\caption{CoP deviation comparison of without SRL support, before optimization and after optimization in WFL task. (a) ML direction. (b) AP direction. $\rm p<0.05$ means there is a significant difference between the second group (Before Optimization) and the third group (After Optimization).}
\label{COP_dev}
\end{figure}

The CoP data in Fig.~\ref{CoP} reveals that, the CoP positions remain within a safe margin both before and after using the SRL, indicating that participants are able to maintain stable walking. 
The area of CoP distribution expands with the SRL support, though a lateral deviation is observed. This deviation is attributed to the added mass on one side, which disrupts the user's mass balance. Despite this, participants are able to control their bodies by swaying within a stable margin. The CoP ellipse area is reduced with the optimized SRL, suggesting improved walking stability. 
However, the analysis of CoP deviation reveals a notable reduction in foot displacement variability with the SRL assistance. 
As shown in Fig.~\ref{COP_dev}, a 16.6\% decrease in CoP fluctuations was observed in the mediolateral (ML) direction (p = 0.0003), whereas the change in the anteroposterior (AP) direction was not statistically significant (p = 0.053). These results suggest that although the impact of SRL is limited in the AP direction, it contributes to improving balance control in the ML direction and enhancing stability during walking tasks.

\begin{figure}[!t]
\centering
\includegraphics[width=1\columnwidth]{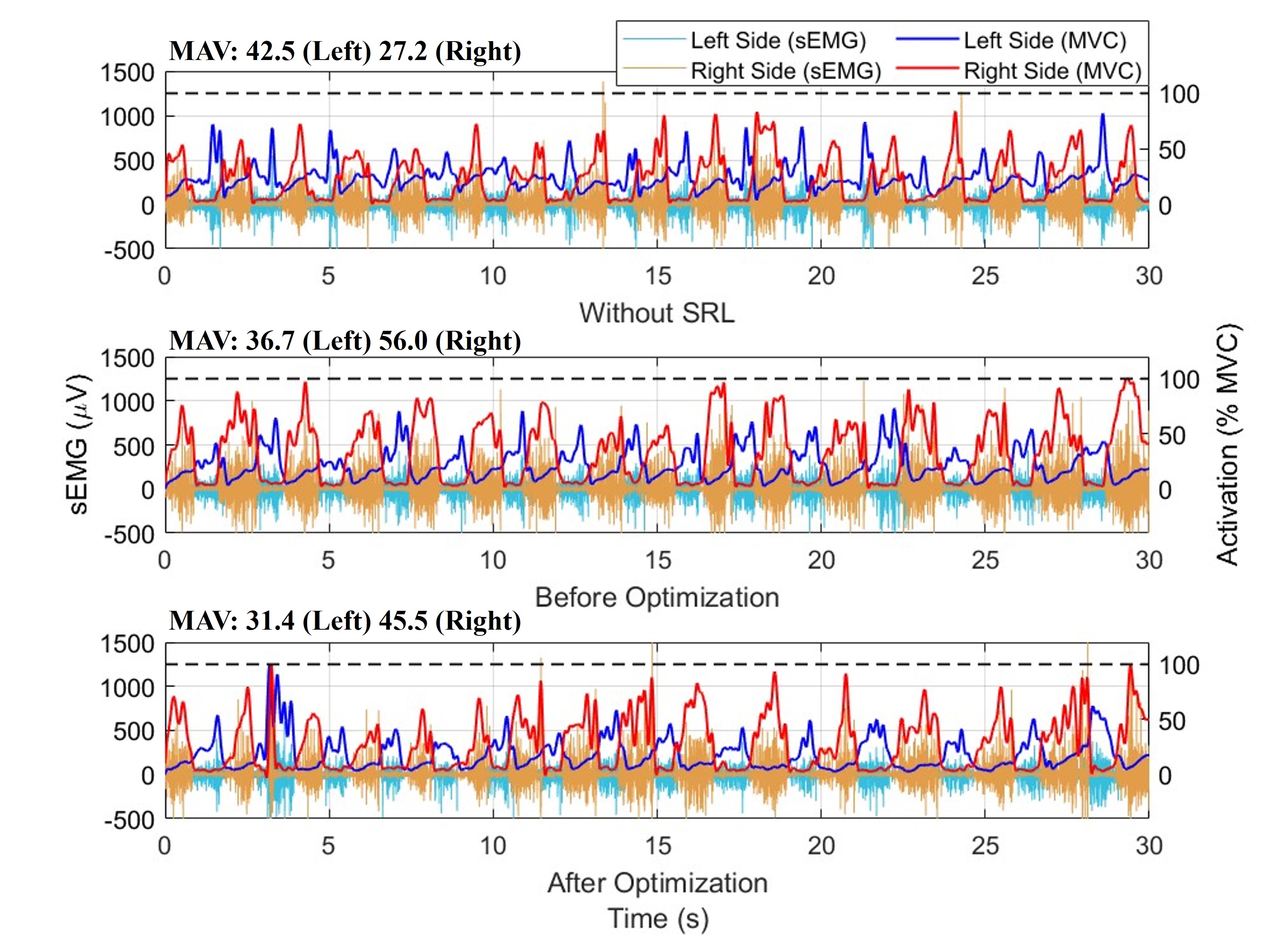}%
\caption{sEMG comparison of without SRL support, before optimization and after optimization in WFL task.}
\label{walkingEMG}
\end{figure}

\begin{figure}[!t]
\centering
\includegraphics[width=1\columnwidth]{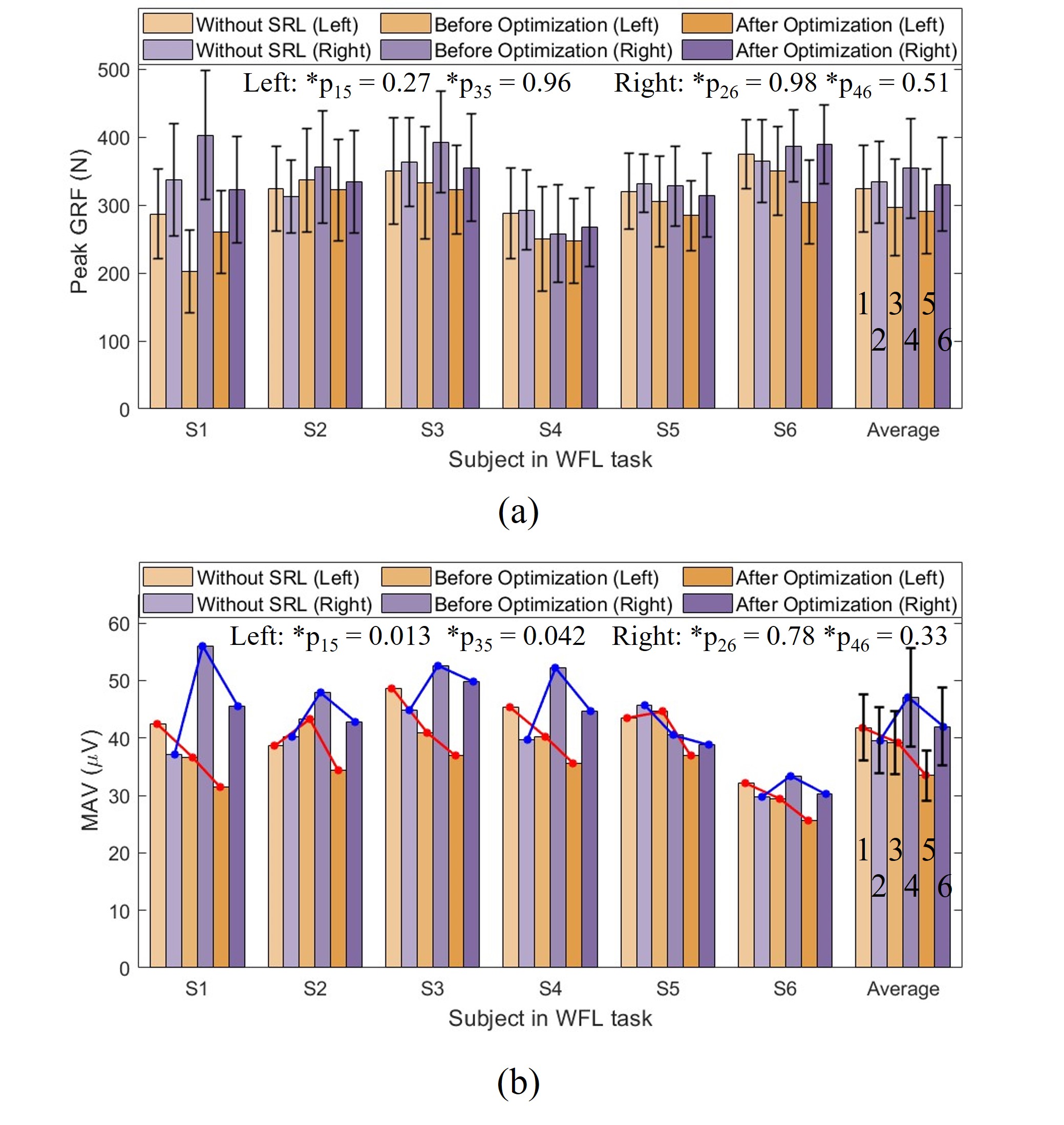}%
\caption{Comparison of subjects support performances in WFL task. The subscripts of the p-value represent the significance comparison of the two columns in the average group. The numbered label is shown in the rightmost column of the figure. For example, $\rm p_{15}<0.05$ means there is a significant difference between the first group (Without SRL, Left Leg) and the fifth group (After Optimization, Left Leg).}
\label{WFL1}
\end{figure}

The peak GRF values during the WFL task (Fig.~\ref{WFL1}(a)) show no significant difference between with and without SRL assistance (p $>$ 0.05).
Fig.~\ref{walkingEMG} and Fig.~\ref{WFL1}(b) illustrate the sEMG comparison of three conditions during the walking task: without assistance, before optimization, and after optimization.
The analysis of sEMG signals reveals that without SRL assistance, muscle activity in both legs is nearly symmetrical. 
With optimized SRL assistance, muscle activity in the left leg decreases significantly by 12.7\% ($\rm p_{35}$ = 0.042) compared to the pre-optimization condition, whereas the increase in the right leg is not statistically significant. This imbalance indicates that the SRL has the ability to reduce muscle activity in the opposite leg.
This phenomenon can be used in various scenarios. For example, for users carrying a unilateral load, placing the SRL on the opposite side can help balance the muscle activity between both legs, thereby enhancing stability. In medical rehabilitation contexts, for hemiplegic patients seeking to reduce fatigue in the affected leg, the SRL can be positioned on the healthy leg. Conversely, if the goal is to train the muscles of the affected leg, the SRL should be placed on the affected side. These findings highlight the potential of the SRL for personalized applications in both daily activities and rehabilitation therapy.

The experiment demonstrates that the universal SRL mechanism proposed in this research has the capability to aid individuals with various walking gaits.
The result of the motion planning is precisely adapted to the natural movement of the human body and the operational task space.

\subsubsection{STS movement}
\begin{figure}[!t]
\centering
\includegraphics[width=0.88\columnwidth]{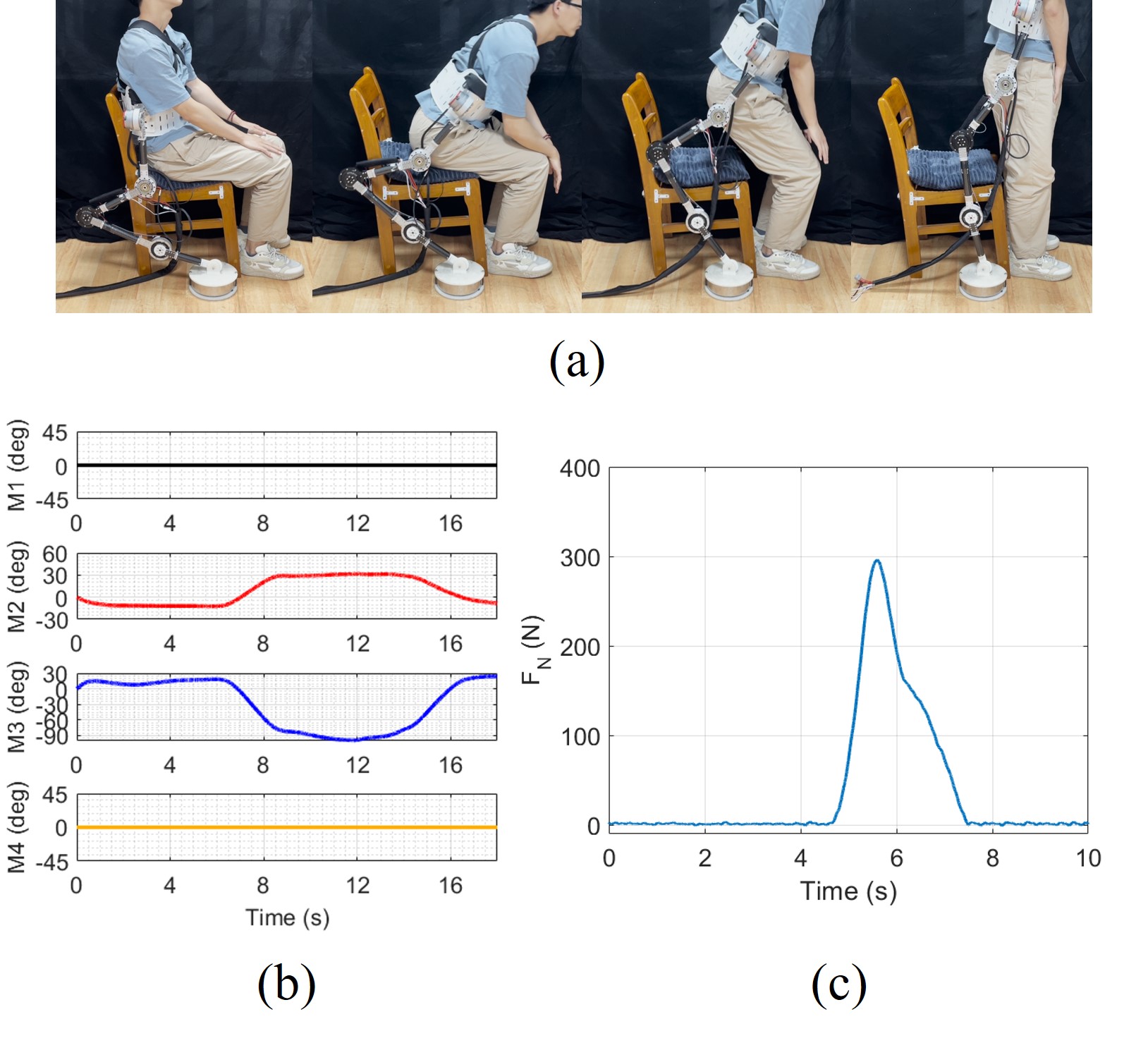}%
\caption{Task 4: STS transfer movement.
(a) Postures sequences demonstrated in STS. 
(b) Joint rotation angle of SRL mechanism. 
(c) The ground reaction force ($F_N$).}
\label{Task5}
\end{figure}

Fig.~\ref{Task5}(a) shows the series of postures conducted during the STS movement experiments.
During the STS transfer function, the proximal and distal joints remain stationary. 
Joint 2 and 3 follow the specified trajectory as shown in Fig.~\ref{Task5}(b).
Fig.~\ref{Task5}(c) shows that the peak ground reaction force (GRF) is approximately 300 N.  
At the onset of the STS task, the mechanism provides maximal support force, which gradually diminishing until reaching zero as the body naturally stands up. 

\begin{figure}[!t]
\centering
\includegraphics[width=1\columnwidth]{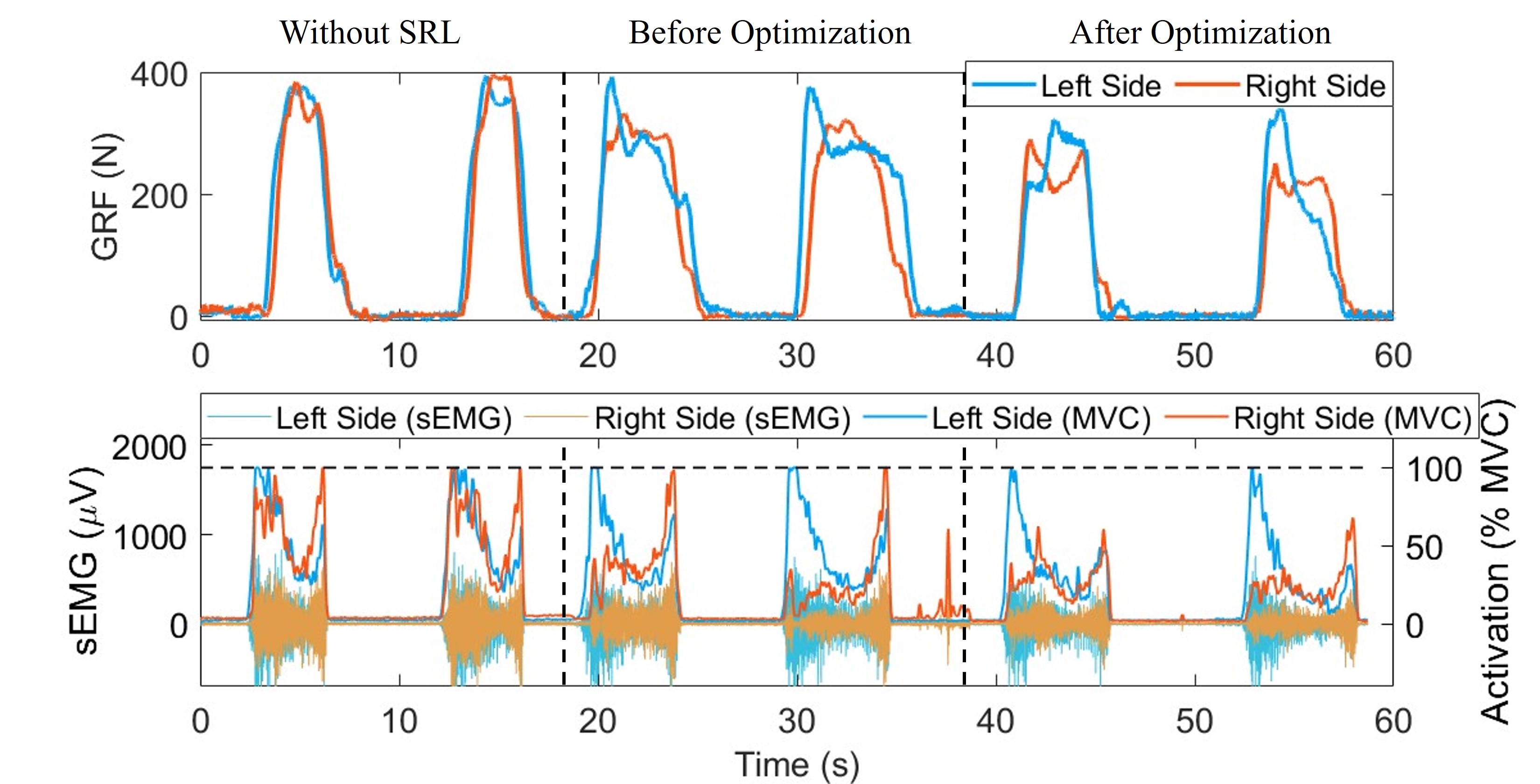}%
\caption{GRF and sEMG comparison of without SRL support, before optimization and after optimization in STS task.}
\label{STSdata}
\end{figure}

\begin{figure}[!t]
\centering
\includegraphics[width=1\columnwidth]{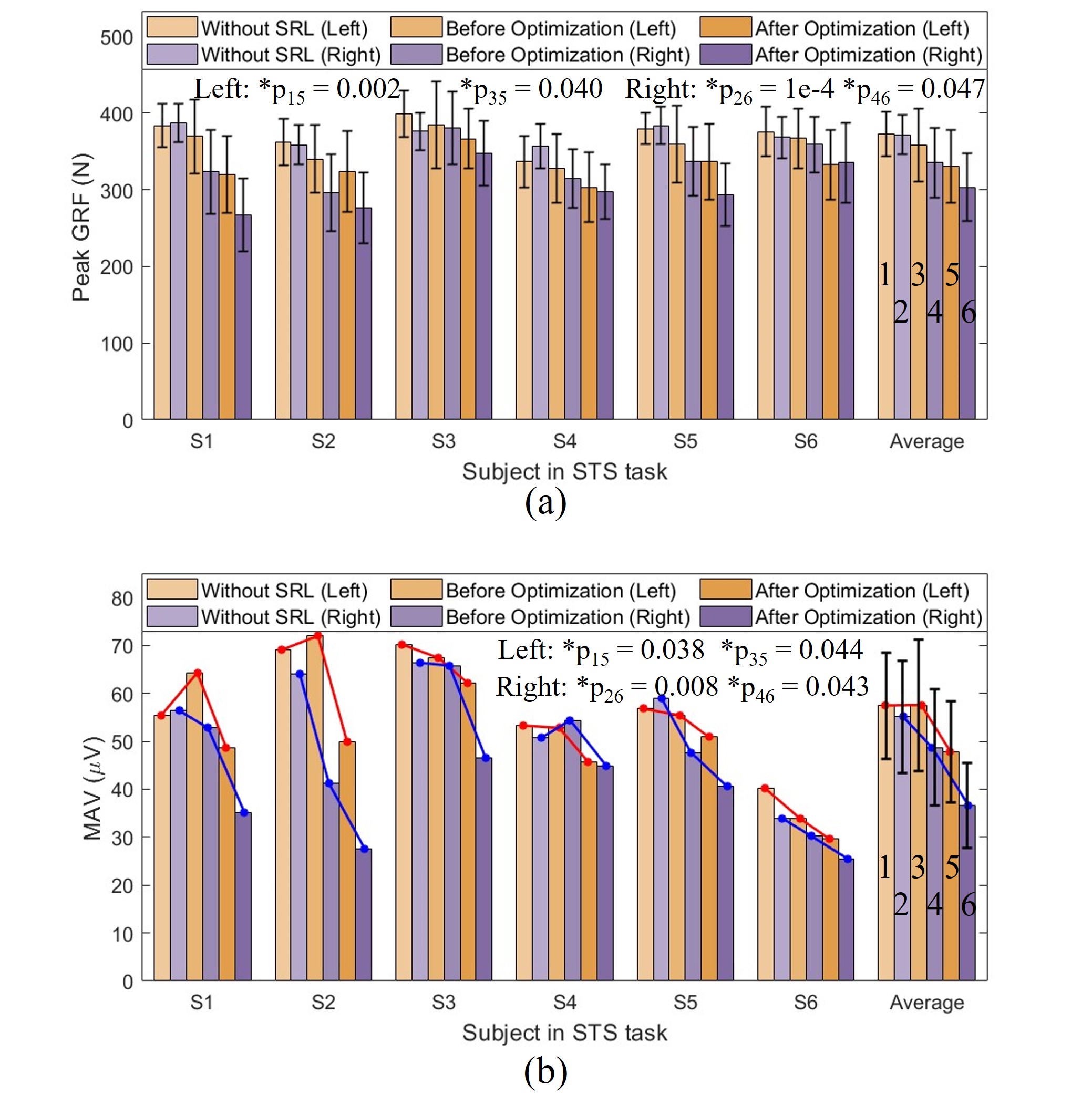}%
\caption{Comparison of participants support performances in STS task. The subscripts of the p-value represent the significance comparison of the two columns in the average group. The numbered label is shown in the rightmost column of the figure. For example, $\rm p_{15}<0.05$ means there is a significant difference between the first group (Without SRL, Left Leg) and the fifth group (After Optimization, Left Leg).}
\label{STSdata2}
\end{figure}

The task is divided into four stages. In the first stage, the participants' upper body was basically vertical to the ground. In the second stage, the subject leaned forward while their lower limb remained on the chair, and the support gradually increased. In the third stage, the subject gradually stood up, and the auxiliary support gradually decreased. In the fourth stage, the subject stood up completely straight, and the support disappeared.

The peak GRF and sEMG indicators are used to assess the performance of the STS motion. 
The vastus lateralis (VL) muscle of the quadriceps was selected as the target muscle due to its effectiveness and ease of electrode placement.
During the experiment, participants were instructed to complete 10 repetitions of the STS motion, with the middle 6 repetitions selected for analysis. sEMG sensors were securely attached to the thigh muscle groups to ensure accurate signal acquisition.

Fig.~\ref{STSdata} and Fig.~\ref{STSdata2} indicate that the SRL effectively reduces the peak GRF and muscle activity during the STS task. 
Compared to the without assistance condition, the optimized SRL reduces peak GRF by an average of 11.3\% for the left leg ($\rm p_{15}$ = 0.002) and 18.5\% for the right leg ($\rm p_{26}$ = 1e-4), muscle activity decreases by 16.7\% for the left leg ($\rm p_{15}$ = 0.038) and 33.4\% for the right leg ($\rm p_{26}$ = 0.008).
The average muscle activity of both legs decreases by 25.1\%. It is also observed that the signals on the right side are consistently weaker than those on the left, reflecting the unilateral assistance provided by the SRL. Additionally, the sEMG signals precede the GRF motion slightly, suggesting that sEMG is a reliable early indicator of STS events. This characteristic makes sEMG a suitable candidate for use as an event trigger in STS tasks. These findings highlight the SRL's potential to assist and optimize the STS process, especially in applications requiring precise event triggering and load reduction.

\subsection{Experiment Validation with Hemiplegia Patients}
To account for the differences between patients and healthy participants, additional experiments were conducted on hemiplegic patients, focusing on the AGO and STS tasks. Tab.~\ref{HS} provides the basic information of the two stroke patients. Prior to the experiments, both patients underwent a Fugl-Meyer limb movement assessment.
Given the limited functionality of their legs, providing walking assistance posed potential safety risks and required more rigorous ethical review. Therefore, experimental verification was confined to the AGO and STS tasks for safety reasons.
\begin{table*}[!t]
\caption{Basic Information of Hemiplegia Subjects}
\label{HS}
\centering
\renewcommand\arraystretch{1.3}
\tabcolsep=0.04\columnwidth
\begin{tabular}{c c c c c c c c}
\hline
\hline
Subject & Gender & Age & Height & Weight & Onset Time & Affected Side & Fugl-Meyer Score\\
\hline
HS1 & Male & 48 & 174 cm & 60 kg & 3.1 year & Left & 34\\
HS2 & Male & 59 & 167 cm & 71 kg & 9.0 year & Left & 44\\
\hline
\hline
\end{tabular}
\end{table*}

In the grasping experiment, the hemiplegic patient initially performed the grasping tasks without SRL assistance. Unlike healthy participants, the patient remained seated due to insufficient lower limb muscle strength. Subsequently, with the assistance of experimenters and physicians, the patient donned the optimized SRL device and completed multiple grasping tasks under the experimenters' guidance. 
Following this, STS experimental trials were conducted on two hemiplegic participants under two conditions: without SRL assistance and with SRL assistance applied to their affected side. The specific experimental data for the two participants are summarized in Tab.~\ref{HS_exp}.
\begin{table}[!t]
\caption{AGO and STS Performance of Stroke Subjects}
\label{HS_exp}
\centering
\renewcommand\arraystretch{1.3}
\tabcolsep=0.04\columnwidth
\begin{threeparttable}
\begin{tabular}{c c c c c }
\hline
\hline
\multirow{2}{*}{Subject} & \multicolumn{2}{c}{Success Rate (AGO)} &
\multicolumn{2}{c}{MAV (STS)}\\
 & WS$^*$ & WOS$^*$ & WS$^*$ & WOS$^*$ \\
\hline
HS1 & 50.0\% & 70.0\% &   74.8 $\mu$V & 72.3$\mu$V\\
HS2 & 40.0\% & 55.0\% &   40.9$\mu$V & 36.2$\mu$V\\
Average & 45.0\% & 62.5\% &  57.9$\mu$V & 54.3$\mu$V \\
\hline
\hline
\end{tabular}
\begin{tablenotes}
\item[*] WS: Without SRL; WOS: With Optimized SRL
\end{tablenotes}
\end{threeparttable}
\end{table}

In the grasping task, without the assistance of the SRL, hemiplegic patients demonstrated significantly lower grasp success rates compared to healthy individuals. This disparity is due to the limited grasping workspace caused by the functional impairment in one arm. However, with SRL assistance, the success rate increased by an average of 17.5\%, highlighting the effectiveness of SRL support. It is worth noting that although the first patient had a lower Fugl-Meyer Score, indicating more severe impairment, their performance in the grasping task was not heavily dependent on the functionality of their arms. 
Instead, 
successful task performance depends more on upper body and abdominal movements, coupled with effective interaction and coordination with the SRL.
In the STS task, compared to un-aided natural STS movements, muscle activity in hemiplegic patients decreases by an average of 6.2\% when assisted by the SRL on the affected side. These patient experiments demonstrate the potential of the proposed SRL for application in hemiplegic rehabilitation scenarios.

\subsection{Discussions}
From a workspace perspective, the objective of this paper is to determine the mechanical configuration that closely replicates the reference workspace. The resulting workspace is expected to match the human workspace in terms of position, shape, and orientation, while also exhibiting an increased size. 
This design theory aims to extend the natural workspace of the human body using the mechanical structure.
In traditional practice \cite{HuoMHS}, the Monte Carlo method is utilized to define the manipulator's workspace using 10,000 discrete 3-D points. 
This phase is computationally intensive, as it requires a substantial number of calculations to analyze the SRL workspace, traverse workspaces, and determine the similarity index.
The algorithmic time complexity is $O(n^2)$.
One advantage of utilizing an ellipsoid to describe the workspace is its ability to describe the workspace through a parameter matrix. The time complexity of the proposed algorithm is $O(n)$, thereby reducing computational complexity. The parameterized ellipsoid workspace provides a more compact and intuitive representation.

The upper limb function requires a gripper for grasping, whereas the lower limb function necessitates an effector with a high friction coefficient for ground contact. Thus, there's a need to replace the end effector with varying functionalities. 
In the STS task, the SRL's endpoint is rigidly fixed to a stationary base plate to ensure stability. Instead, the WFL task employs a mobile end contact configuration.  A reconfigurable end-effector could be designed to enable smooth transitions between different operational tasks in future work.
Three different SRL assisted walking gaits were implemented in \cite{Khazoom}. In this research, SRL assisted walking is implemented with unilateral human limb synchronization, without introducing the third-leg gait, which involves a half-cycle phase difference. This is due to the fact that the three gaits share an identical workspace and differ only in their phase. We did not replicate the third-leg gait as it does not represent a major innovation and contribution in this research.

\begin{table*}[!t]
\caption{Comparison with Other SRL Robot}
\label{MassComparison}
\centering
\renewcommand\arraystretch{1.3}
\tabcolsep=0.04\columnwidth
\begin{tabular}{c c c c c}
\hline
\hline
Research & DoFs/No. of limbs & Installation location & Functionality & Mass \\
\hline
Bonilla et al. \cite{Bonilla2014}, 2014 & 5/2 & Shoulders & Overhead task & 8 kg\\
Parietti et al.\cite{PariettiTRO}, 2016  & 3/2 & Hip bone & Balancing and stabilizing & 13 kg \\
Gonzalez et al. \cite{GonzalezRAL}, 2019 & 3/2 & Back & Balancing and stabilizing & 9.6 kg\\
Véronneau et al.\cite{Veronneau}, 2020 & 3/1 & Waist & Grasping assisting & 7.2 kg \\
Khazoom et al.\cite{Khazoom}, 2020 & 2/1 & Hip bone &  Walking assisting & 9.7 kg \\
Hao et al. \cite{Fu2020JMR}, 2020 & 4/2 & Back & Walking assisting & 6.5 kg \\
Zhang et al. \cite{ZhuYanheAIS}, 2024 & 6/2 & Shoulders & Locating assistance & 6.7 kg \\
Abeywardena et al. \cite{Abeywardena_2024}, 2024 & 2/1 & Pelvis & Balancing tail & 10 kg\\
Luo et al. \cite{LuoTSMC}, 2025 & 6/1 & Waist & Dynamic grasping balancing & About 5 kg \\
Zhang et al.\cite{SongAiguoTASE}, 2025 & 3/1 & Shoulders & Overhead task &2.25 kg\\
Proposed & 4/1 & Waist & Grasping and walking assisting & 5.0 kg \\
\hline
\hline
\end{tabular}
\end{table*}

In this study, the link length is optimized as a key design parameter. Other parameters, such as link twist and link offset, also affect the mechanism's workspace. 
These parameters are excluded for two reasons.
First, to replicate the joint configuration of human upper and lower limbs, we constrain the lower limb's motion to the Y-Z plane. This constraint limits the effect of link twist on the workspace.  Previous research \cite{HuoMHS} has shown that the optimal link twist matches the current configuration. Additionally, including link offset may complicate the design, increase mass and inertia and compromise the goal of creating a lightweight prosthetic.
Second, in multi-objective optimization, adding variables increases computational complexity and makes it more challenging to find an optimal solution. Therefore, we focus exclusively on optimizing link length.

In the optimization framework presented in this paper, the mechanical structure constraints restrict the connecting rod from achieving a match with the optimal solution. For instance, although the preferred length for the initial connecting rod near the user is 0.1 m, practical considerations require an adjustment to 0.15 m to accommodate the connector's length and motor dimensions. To retain the overall length, the second link's length is reduced by 0.05 m accordingly.
Additionally, the effective total length of the mechanism is designed based on the subject's height, particularly focusing on the length of the lower limbs. 
It should be noted that this optimization method is custom-designed for the user. Differences in leg length among users may yield different optimization results.
Therefore, the proposed mechanism exhibits limited adaptability to various subjects.

We have provided a comprehensive comparison in Table \ref{MassComparison} that highlights the  differences in  key configurations between our SRL and other benchmark systems, including the number of DoFs (per limb) and limbs, installation location, functional capabilities, and overall mass.
SRLs designed for upper-limb tasks are typically mounted on the shoulders or waist, while those intended for lower-limb tasks are generally mounted at the pelvis. Given that our SRL is designed to support both upper and lower limb functions, a waist-mounted configuration is selected to accommodate this dual functionality. 
While extant SRL research primarily focuses on single-purpose devices, our proposed SRL integrates multiple functions into a compact and lightweight structure, offering distinct advantages in versatility and usability.

Experimental observations and user feedback indicate the presence of certain interference effects on natural arm swing dynamics during SRL operation. However, for our primary target population, hemiplegic patients, the impact is clinically negligible. Due to the paralyzed state of the affected limb, its contribution to functional mobility is inherently minimal. As a result, the usability of the unaffected upper limb is largely preserved. 
For healthy users, their upper limb movements are sensitive to interference, and some participants report noticeable changes in their gait patterns and lower limb movements during prolonged use.
This may be related to occasional interactions between the natural arm swing during walking and the movements of the SRL.
Specifically, we observed that although the SRL effectively aids in load redistribution and enhances gait stability (see Figure \ref{COP_dev} - \ref{WFL1}), its mass and asymmetrical mounting position may induce compensatory movements in the torso or contralateral limb during prolonged use. Such compensations may lead to user fatigue or long-term alterations in gait patterns, warranting further investigation.
Moreover, this study highlights open questions regarding long-term biomechanical adaptation and individual variability.
Future work should incorporate musculoskeletal modeling and sEMG-based feedback to more accurately quantify the biomechanical cost and neuromuscular responses with SRL-assisted walking.

When used for upper limb functions, the SRL can support more complex tasks, such as bi-manual or tri-manual operations. 
In these tasks, it is important to ensure that the SRL’s workspace sufficiently overlaps with the human limb’s workspace to maximize functionality. The adopted workspace metrics can capture the functional requirements of the tasks to some extent. Additionally, a video is included in the supplementary material to demonstrate a bi-manual collaboration experiment.
The tasks in the collaboration experiment require advanced kinematic and dynamic analysis to support effective human–robot perception and collaboration between the user and the SRL. The perceptual requirements differ between agents: humans must generate motion and intention signals, whereas robotic systems require integrated multi-modal perception encompassing visual, haptic, and acoustic channels. Building on this perceptual interaction, different control paradigms can be implemented, such as human-dominant with robot assistance, robot-dominant with human assistance, and shared control \cite{Yang2020THMS}. 
This area is a promising avenue for future research.

In practical wearable robotic systems, the robot base is often movable, resulting in a floating-base robot system. The control of such floating-base systems relies on feedback from the current system state to estimate the disturbance caused by the movement of the floating base on the SRL system. Therefore, developing a comprehensive dynamics model is essential and challenging. The strong coupling between the base and the SRL mechanism complicates the modeling process, as the strong interdependence inherent in the floating system increases the overall complexity of the system's dynamics. Therefore, establishing a coupled dynamic model of the base-SRL system within the floating-base framework is of critical importance. The control of the floating-base robot system is the focus of our future research.

In rehabilitation medicine, gaits are commonly categorized into two-point  and three-point patterns \cite{RASOULI2020109489}. These gait patterns are widely adopted for crutch-assisted walking and are well-accepted and adaptable for most users.
For healthy individuals and hemiplegia patients, the required gait assistance patterns are different. Two-point gait proves effective for partial weight-bearers, including healthy individuals and patients with moderate walking ability. The three-point gait is intended for individuals who are partially or completely unable to bear weight on one leg, while retaining normal muscle strength in the other. As such, it is particularly suitable for patients with unilateral lower limb weakness.
Currently, two-point gait based on offline-recorded motion trajectory is used to generate the SRL motion trajectory, which limits its adaptability and real-time applicability. The requirements for real-time trajectory generation vary across tasks and users. 
In grasping tasks involving wearable robots with movable bases, precise localization of both the target object and the user's posture presents a significant challenge for perception and control systems. 
In walking tasks, current trajectory generation approaches mainly involve deriving gait trajectories from the healthy limb to replicate natural movement patterns \cite{CLME}, or adjusting gait phases to achieve asynchronous gait coordination with the user's limb \cite{Khazoom}. Given the considerable variability in user gait patterns, cross-user gait generation may benefit from a transfer learning approach. 
For STS tasks, quick-response real-time control is essential due to its rapid execution. 
Given their anticipatory nature \cite{Capotorti2022sEMG}, sEMG signals serve as effective triggers for initiating STS tasks. These challenges will be further addressed in our future work.

The biomechanical response of the human body must be considered at both the control and design levels.
One effective approach for assessing the biomechanical impact of wearable robotic systems is the Computed Muscle Control (CMC) method, implemented via OpenSim software. This framework enables the development of musculoskeletal models and the simulation of muscle forces under varying mechanical parameters and motion trajectories \cite{VatsalRAL}. Using Hill-type muscle-tendon units, CMC can estimate muscle activation and force output, thereby allowing for the computation of metabolic cost and physiological effort \cite{Abeywardena_2024,Uchidapone,StingelRAL}. As this method relies on prior knowledge of motion trajectories, it is particularly well suited for analyzing the dynamic behavior of human-robot interactions.

Predictive modeling with OpenSim should be considered not only as a post-hoc analysis tool, but also as a proactive component during the design phase of SRLs and other wearable devices. Integrating biomechanical performance indicators derived from simulation into the optimization loop can enhance the comprehensiveness and reliability of the design framework. The future of wearable robotics, particularly the development of SRL, will benefit from the integration of the optimization and control methodologies proposed in this work with the simulation-based design approaches used in musculoskeletal research. This fusion facilitates the prediction of biomechanical effects resulting from design choices and allows for refinement prior to physical prototyping.

\section{Conclusion}
This study proposes a multi-objective optimization approach to improve the parameters of a general-purpose supernumerary robotic limb, which is designed to enhance the human limb ability. It focuses on combining the upper limb grasping workspace and lower limb force output to provide a comprehensive solution. 
An approach using ellipsoids for geometric quantification is described to address the challenges related to computing workspace. The ellipsoid envelope is used to convert workspace features into ellipsoid parameters, allowing for a parametric representation of the workspace. A workspace similarity index is also introduced. 
The work addresses difficulties in achieving stable convergence to the high-dimensional irregular pareto front in multi-objective optimization by developing a model using performance indicators. 
The model's optimal solution enables redesign of the SRL prototype platform to meet specific requirements. 
The proposed framework offers a robust model for designing multi-functional SRL mechanisms, showing great potential in rehabilitating patients with hemiplegia and enhancing the physical strength of healthy individuals.

The proposed SRL mechanism is validated through tests involving healthy participants and patients with hemiplegia in the present study. Future studies may include a larger cohort of patients to further evaluate the performance of the proposed framework.

\bibliographystyle{IEEEtran}
\bibliography{Refs}





\vspace{11pt}


\begin{IEEEbiography}[{\includegraphics[width=1in,height=1.25in,clip,keepaspectratio]{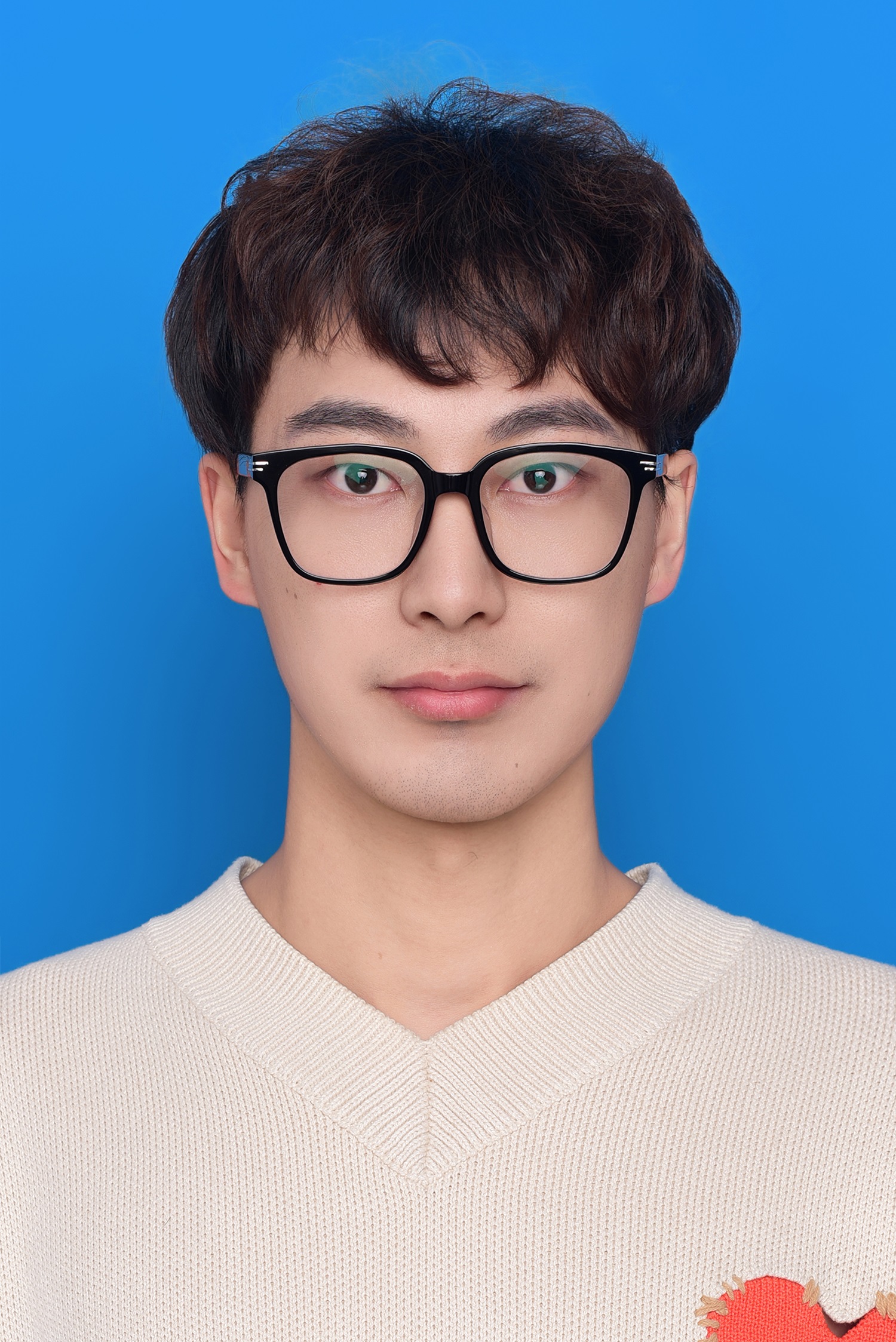}}]{Jun Huo}
received the B.S. degree in mechanical engineering from Northeastern University, Shenyang, China, in 2018. He is currently working towards the Ph.D. degree in control science and engineering of Huazhong University of Science and Technology. His current research interests include mechanical design, sensing and control of rehabilitation assistive robotic systems (wearable robots). 
\end{IEEEbiography}

\begin{IEEEbiography}[{\includegraphics[width=1in,height=1.25in,clip,keepaspectratio]{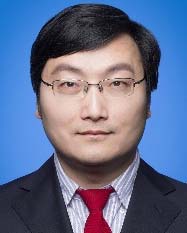}}]{Jian Huang}
(Senior Member, IEEE) received the bachelor’s degree in automatic control, the M.E. degree in control theory and control engineering, and the Ph.D. degree in control science and engineering from the Huazhong University of Science and Technology (HUST), Wuhan, China, in 1997, 2000, and 2005, respectively.

From 2006 to 2008, he was a Postdoctoral Researcher with the Department of Micro-
Nano System Engineering and the Department of Mechano-Informatics and Systems, Nagoya University, Nagoya, Japan. He is currently a Full Professor with the School of Artificial Intelligence and Automation, HUST. His main research interests include rehabilitation robots, robotic assembly, networked control systems, and bioinformatics.
\end{IEEEbiography}

\begin{IEEEbiography}[{\includegraphics[width=1in,height=1.25in,clip,keepaspectratio]{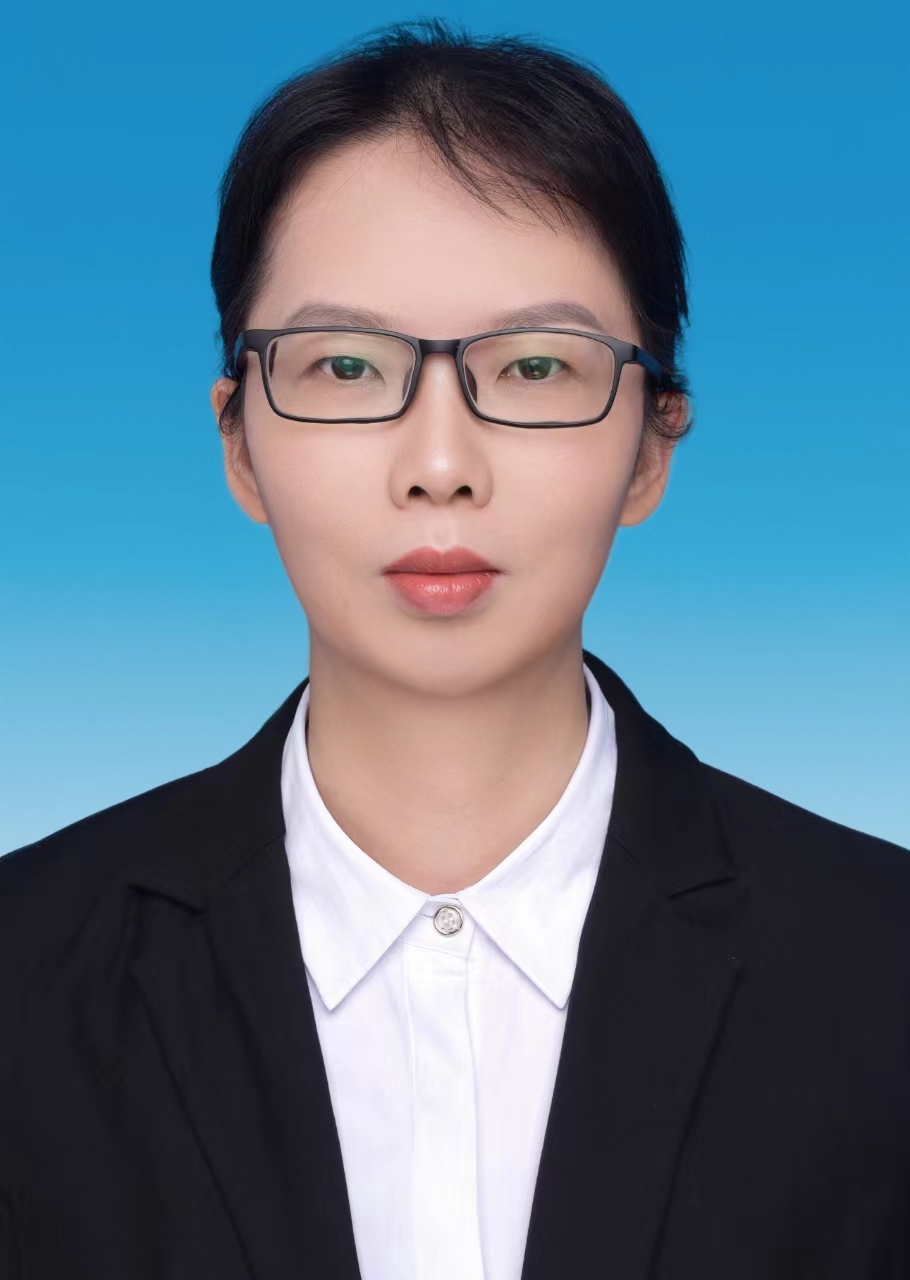}}]{Jie Zuo}
received the Ph.D. degree in information engineering from Wuhan University of Technology, Wuhan, China, in 2022. She was a postdoctoral researcher in the School of Artificial Intelligence and Automation at Huazhong University of Science and Technology, and is currently working in the School of Information Engineering at Wuhan University of Technology. Her research interests include rehabilitation robots modelling and control strategies.
\end{IEEEbiography}

\begin{IEEEbiography}[{\includegraphics[width=1in,height=1.25in,clip,keepaspectratio]{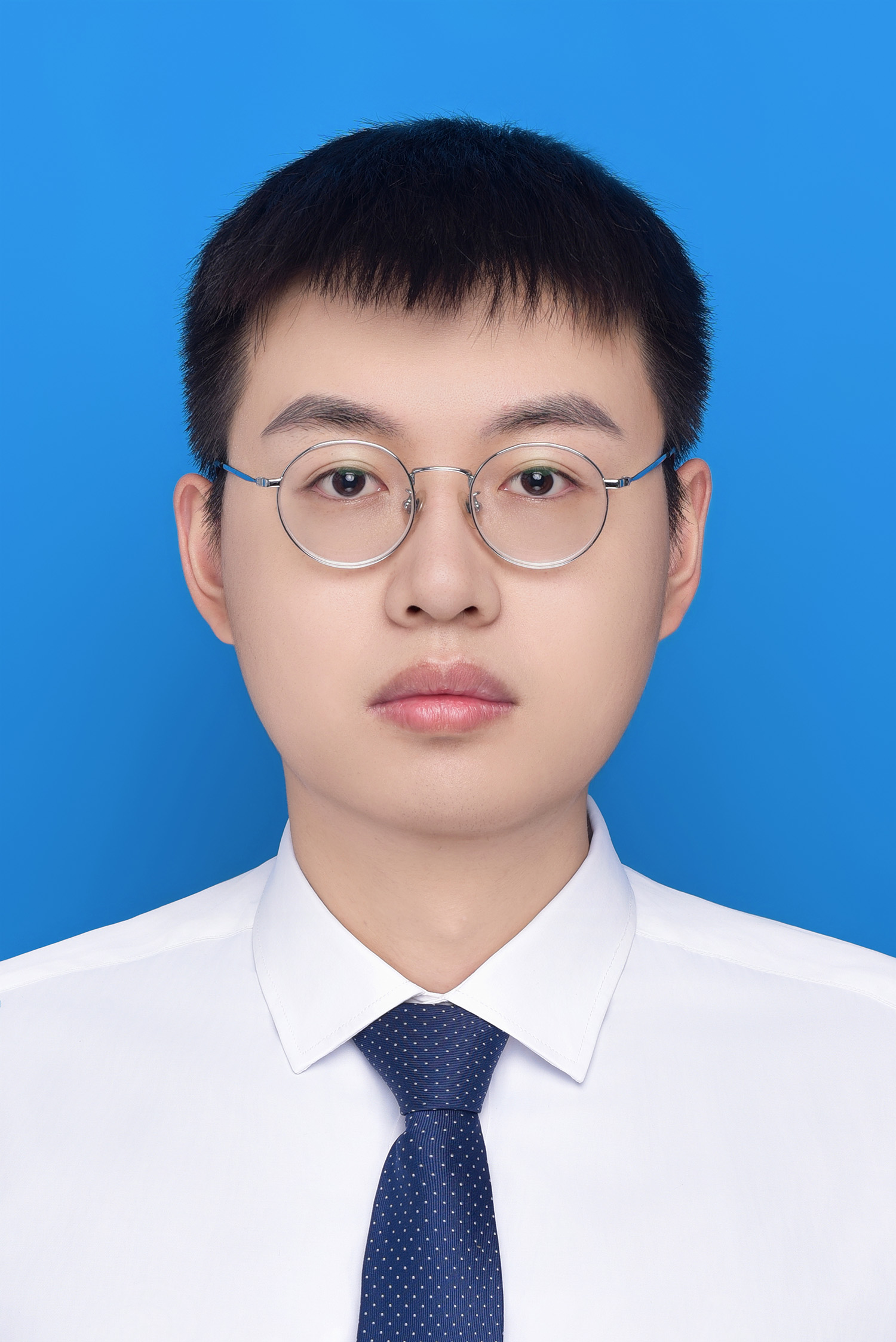}}]{Bo Yang}
received the B.S. degree in automation from Chongqing University, Chongqing, China, in 2019, and the Ph.D. degree in control science and engineering of Huazhong University of Science and Technology, Wuhan, China, in 2024. His current research interests include human intention recognition, intention-based robot control, and set-membership filter.
\end{IEEEbiography}

\begin{IEEEbiography}[{\includegraphics[width=1in,height=1.25in,clip,keepaspectratio]{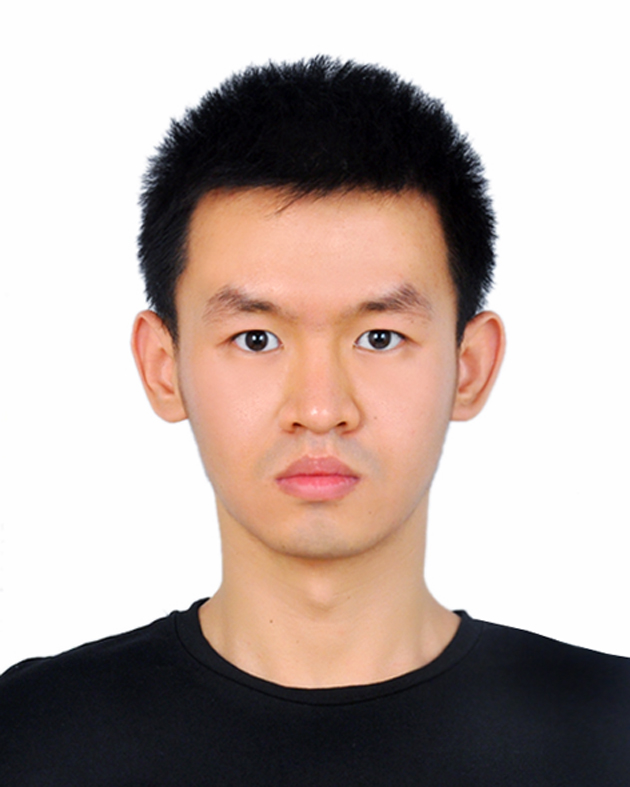}}]{Zhongzheng Fu}
is currently a doctoral student in Man-Machine Fusion and Artificial Intelligence Application, School of Artificial Intelligence and Automation, Huazhong University of Science and Technology in China. His main fields of research include transfer learning, affective computing, gesture recognition and application of evolutionary algorithms in optimization.
\end{IEEEbiography}

\begin{IEEEbiography}[{\includegraphics[width=1in,height=1.25in,clip,keepaspectratio]{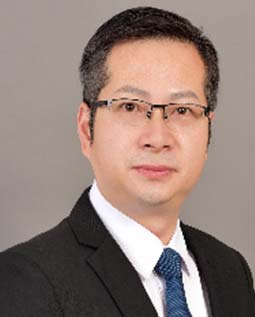}}]{Xi Li}
received the Ph.D. degree from Shanghai Jiao Tong University, Shanghai, China, in 2006.
From 2011 to 2012, he was a Visiting Scholar with the Naval Architecture and Marine Engineering Department, University of Michigan, Ann Arbor, MI, USA. He has been with the Huazhong University of Science and Technology, Wuhan, China, since 2006, where he is currently a Professor with the School of Artificial Intelligence and Automation. His research interests include intelligent control and model-predictive control of renewable energy systems.
\end{IEEEbiography}

\begin{IEEEbiography}[{\includegraphics[width=1in,height=1.25in,clip,keepaspectratio]{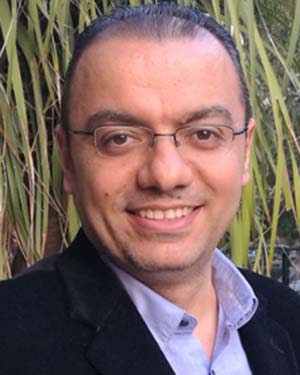}}]{Samer Mohammed}
(Senior Member, IEEE) received the Ph.D. degree in computer science from the Laboratory of Computer Science, Robotic and Microelectronic of Montpellier (LIRMM/CNRS), Montpellier and the Habilitation degree in robotics from the University of Paris-Est Créteil, in 2016. He is currently Full Professor with the Laboratory of Images, Signals and Intelligent Systems, University of Paris-Est Créteil, Créteil, France. He has authored or co-authored more than 100 papers in scientific journals, books, and conference proceedings. His current research interests include modeling, identification, and control of robotic systems (wearable robots), artificial intelligence, and decision-support theory. Target applications concern mainly the functional assisting of dependent people. 

Dr. Mohammed was the general chair of the IEEE-RAS Technical Committee on Wearable Robotics. He was the recipient of the Fellowship Award from the Japan Society for the Promotion of Science and the French Ministerial Doctoral Research Scholarship Award. He is actively involved in different national and international projects. He co-organized several national and international workshops and conferences in the field of wearable technologies.



\end{IEEEbiography}



\vfill

\end{document}